\documentclass{article}

\usepackage{arxiv}

\usepackage[utf8]{inputenc} 
\usepackage[T1]{fontenc}    
\usepackage{hyperref}       
\usepackage{url}            
\usepackage{booktabs}       
\usepackage{multirow}
\usepackage{amsmath}        
\usepackage{amssymb}        
\usepackage{amsfonts}       
\usepackage{nicefrac}       
\usepackage{microtype}      
\usepackage{graphicx}
\usepackage{natbib}
\usepackage{doi}
\usepackage{xcolor}

\definecolor{genablue}{RGB}{0,90,200}

\definecolor{soslangreen}{RGB}{0,130,70}

\definecolor{maksimpurple}{RGB}{140,40,170}

\definecolor{wolforange}{RGB}{200,90,0}

\newcommand{\gain}[1]{{\scriptsize\textcolor{gray}{#1}}}

\newcommand{\tabnote}[1]{
  \par\vspace{3pt}
  \begin{minipage}{\linewidth}\footnotesize\raggedright #1\end{minipage}}

\title{CADENA: Stepwise CAD Reverse Engineering}

\author{
\mdseries Soslan Kabisov\textsuperscript{1}\textsuperscript{*}\\
\and
Gennadiy Savrasov\textsuperscript{1,6}\textsuperscript{*}\\
\and
Maksim Elistratov \textsuperscript{1}\\
\and
Antonio Rodriguez \textsuperscript{2}\\
\and
Daniil Ignatiev \textsuperscript{2} \\
\and
Nikita Gavrilov \textsuperscript{1} \\
\and
Rustam Uzdenov \textsuperscript{1}\\
\and
Alexey I. Boyko \textsuperscript{3}\\
\and
Igor Pasechnik  \textsuperscript{2}\\
\and
Anton Konushin \textsuperscript{1,6}\\
\and
Andrey Kuznetsov \textsuperscript{4}\textsuperscript{5}
\\
\and
Dmitrii Zhemchuzhnikov\textsuperscript{1,6}\dag
\and
\\
\textsuperscript{1} Lomonosov Moscow State University;
\textsuperscript{2} DAIMLD; \\
\textsuperscript{3} Independent Researcher; \\
\textsuperscript{4} Innopolis University;
\textsuperscript{5} FusionBrain Lab; \\
\textsuperscript{6} AXXX, Moscow, Russia
}

\renewcommand{\shorttitle}{CADENA: Stepwise CAD Reverse Engineering}

\hypersetup{
pdftitle={CADENA: Stepwise CAD Reverse Engineering},
pdfsubject={cs.CV, cs.LG},
pdfauthor={Soslan Kabisov, Gennadiy Savrasov, Maksim Elistratov, Antonio Rodriguez, Daniil Ignatiev, Nikita Gavrilov, Rustam Uzdenov, Alexey I. Boyko, Igor Pasechnik, Anton Konushin, Andrey Kuznetsov, Dmitrii Zhemchuzhnikov},
pdfkeywords={CAD reverse engineering, parametric CAD, vision-language models, CadQuery},
}

\begin{document}
\maketitle
\renewcommand{\thefootnote}{\fnsymbol{footnote}}
\footnotetext[1]{Equal contribution.}
\footnotetext[2]{Corresponding author: zhemchuzhnikovds@my.msu.ru}
\renewcommand{\thefootnote}{\arabic{footnote}}

\begin{abstract}
Computer-Aided Design (CAD) underpins modern engineering, yet converting existing shapes into editable models still demands substantial expert effort. Most AI systems emit the entire CAD program in a single pass, never inspecting the intermediate geometry. In contrast, human engineers build a part feature by feature, checking after each operation what remains to be modeled. We introduce CADENA (Spanish for ``chain''), a model that reconstructs a 3D mesh as a parametric CAD program, growing its sequence of operations one at a time and comparing the target with the currently predicted geometry at every step. We also address the lack of benchmarks for evaluating reverse-engineering methods on mechanical parts, introducing CADENA-Bench, a benchmark that measures performance across categories of mechanical parts. CADENA outperforms prior methods on CADENA-Bench and on the DeepCAD, Fusion 360, and MCB datasets.
Code is available at \url{https://github.com/zhemdi/cadena}, model weights at \url{https://huggingface.co/kulibinai/cadena}, and CADENA-Bench at \url{https://huggingface.co/datasets/kulibinai/cadena-bench}.
\end{abstract}

\keywords{CAD reverse engineering \and parametric CAD \and vision-language models}

\section{Introduction}
\label{sec:intro}

Editable parametric CAD models are the working currency of mechanical engineering, yet most existing 3D assets --- scanned parts, legacy archives, models exported without history --- exist only as raw geometry. A mesh can be rendered and printed but not changed: it has no notion of a bore, a fillet or a wall, only triangles, so there is no handle by which to widen a hole or lengthen a bracket. A part is almost never reused exactly as found --- it must be adapted to the standards of the product it goes into --- and every such adaptation is an edit to a parameter that only a program exposes. The case is stronger still for scanned input, whose surfaces are noisy and edges rounded: fitting geometry to such a measurement reproduces its defects, whereas recovering a program reconstructs what the part \emph{is} --- a cylinder of some diameter, a hole at some position --- and with it idealised, manufacturable geometry.

Learning-based reverse engineering has made this tractable: vision--language models translate renders or point clouds directly into executable CAD code. But these systems emit the whole program in one pass and never inspect the geometry their code produces, and two failures follow. The model commits to each operation without seeing what its earlier ones built, so an early mistake compounds silently; and it conditions on the program text written so far rather than on the geometry that remains, so its choice follows the compositional statistics of the training corpus rather than the shape in front of it. Conditioning on the residual --- what the target still has that the build does not --- removes both, and any operation may be applied at any point, because nothing about the choice depends on how far along the program is. CADENA closes this loop, emitting one operation at a time and executing it before deciding the next (Fig.~\ref{fig:method}). What makes this practical is a property single-pass methods leave unused: the target shape is the \emph{input} to reverse engineering, not a hidden label, so it is available at inference as well as in training. 

Evaluating such systems on real mechanical parts is its own unsolved problem. Existing test sets are dominated by simple sketch--extrude shapes; corpora of real mechanical parts do exist, but they were assembled for classification and retrieval rather than reconstruction, and consist largely of simple primitives and of near-duplicates, both within and across sources. The standard metrics are also poorly suited to it: they score the recovered shape rather than the recovered program, so a body assembled from the wrong primitives can satisfy them. We therefore assemble CADENA-Bench from three such corpora, discarding trivial and duplicate parts and grouping the 3396 that remain into six families by the features from which they are built, and evaluate with the \emph{Generalized Match Score} (GMS). GMS scores agreement of surface type rather than of occupied volume, and is defined for parts that are not watertight solids.

Our contributions are:
\begin{itemize}
    \item CADENA, a stepwise reverse-engineering model that grows a CAD program one operation at a time under explicit geometric feedback, trained by supervised fine-tuning and refined with online reinforcement learning rewarded by executed geometry, and reaching the best reported results on all five datasets we evaluate --- DeepCAD, Fusion360, MCB, CADENA-Bench and BenchCAD (Tables~\ref{tab:external},~\ref{tab:benchmark} and~\ref{tab:benchcad});
    \item CADENA-Bench, a benchmark of 3396 real mechanical parts in six part families, reported per family so that a method's domain of applicability is visible rather than averaged away;
    \item GMS, a surface-matching metric.
\end{itemize}

\section{Related Work}
\label{sec:related}

\paragraph{Generative CAD modelling.} Sequence models over CAD construction histories generate programs rather than raw geometry, but they target unconditional or text-conditioned synthesis rather than reconstruction of a given shape, and their representations are largely restricted to sketch-and-extrude. DeepCAD~\citep{deepcad} models sketch--extrude sequences with a transformer and contributed the corpus most subsequent work is trained on; the Fusion 360 Gallery~\citep{fusion360} provides human design sequences together with an environment that exposes construction as a Markov decision process; SkexGen~\citep{skexgen} disentangles topology, geometry and extrusion into separate codebooks for controllable synthesis. Because sketches are the atoms of these sequences, a parallel literature models them directly, with autoregressive~\citep{sketchgen, vitruvion}, multimodal~\citep{cadvlm}, and diffusion~\citep{sketchdnn} formulations that emit primitives together with the constraints linking them; more recent systems condition on several modalities at once~\citep{cadmllm} or adapt tokenisation to CAD's primitive structure~\citep{cadtokenizer, hiercad}. Recent work pushes on the sketch--extrude restriction from two sides. From the data side, CADFS~\citep{cadfs} adopts a FeatureScript representation covering fifteen operations, Zero-to-CAD~\citep{zerotocad} synthesises a million executable sequences through agentic search, and CADEvolve~\citep{cadevolve} evolves programs from primitives toward industrial complexity. From the representation side, Pointer-CAD~\citep{pointercad} adds explicit B-Rep entity selection so that operations such as fillet and chamfer can be expressed at all, and HistCAD~\citep{histcad} records the constraints that make edits propagate. CADENA takes the data problem in the same spirit, but for \emph{stepwise} supervision: its generator emits not only programs but the intermediate states needed to learn one operation at a time.

\paragraph{Reconstruction into non-program representations.} A large body of work reverse-engineers geometry into representations that are structured but not executable programs. One line recovers boundary representations directly, either by detecting and assembling primitives~\citep{complexgen, splitandfit} or by generating B-Reps with autoregressive and diffusion models~\citep{solidgen, brepgen, hola, autobrep, brepgpt, paracad, dualbrep, brdf}. Another decomposes shapes into constructive solid geometry or primitive assemblies, supervised~\citep{csgnet} or unsupervised~\citep{ucsgnet, caprinet, d2csg, resfit}. A third targets the sketch--extrude subset specifically, inferring extrusion cylinders~\citep{point2cyl}, learning inverse sketch-and-extrude without supervision~\citep{extrudenet, secadnet}, or searching over modelling sequences with zone graphs~\citep{zonegraphs}. These methods produce editable geometry, but not the construction history in a language an engineer can read and modify; our target is the program itself.

\paragraph{CAD reverse engineering.} The systems closest to ours map geometry directly to executable programs, and what separates them is how often the program is executed while it is being written. In most of them, never: CAD-Recode~\citep{cadrecode} maps point clouds to CadQuery programs with an LLM decoder; cadrille~\citep{cadrille} accepts point clouds, images, and text, and is the first to apply online RL fine-tuning to the task; CADEvolve~\citep{cadevolve} scales supervision through evolved training programs; and vision--language models fine-tuned to emit CadQuery from images follow the same pattern~\citep{cadcodervlm}. All of these emit the entire program in a single pass. CADFit~\citep{cadfit} is at the other extreme, but is not a learned generator: it reconstructs by optimisation, incrementally fitting and validating parametric operations against the target under an IoU objective and pruning the candidate set with a learned sketch prior, so its search cost grows with part complexity --- a trade-off its results make visible (Section~\ref{sec:results}). CADReasoner~\citep{cadreasoner}, our closest prior work, executes in between: it closes the loop by re-generating the \emph{whole} program across refinement rounds conditioned on the discrepancy between input and current reconstruction. CADENA differs in granularity: rather than editing a complete program, it appends one operation at a time, so each decision is made against the geometry built so far and no decision is conditioned on the program text.

\paragraph{Iterative and stepwise generation.} A parallel line of work replaces single-pass decoding with closed loops, but in almost all of it the unit of interaction is a whole program and the feedback is mediated by language. Agent-style systems drive frozen or fine-tuned VLMs through multi-turn interaction with a CAD sandbox: CAD-Assistant~\citep{cadassistant} executes actions against a CAD kernel through its Python API and adapts to the evolving design state, IterCAD~\citep{itercad} formulates generation and editing as multi-turn interaction with an executable sandbox, and ToolCAD~\citep{toolcad} and COSMO-Agent~\citep{cosmoagent} train LLMs as tool-users over CAD and simulation engines. Related efforts use execution as a training signal rather than an inference loop, rewarding geometric agreement~\citep{cadcodercot, cmecad} or compiler feedback~\citep{cadjudge}, and a further line targets editing of existing models~\citep{cadeditor, prcad}.

In the work SOV-CAD~\citep{sovcad}, released while this work was in preparation, similar to our approach, CAD reconstruction is performed iteratively using visual feedback obtained by rendering the partially constructed model together with the target. However, the methodological foundations differ substantially. SOV-CAD formulates reconstruction as an offline reinforcement learning problem using a Decision Transformer operating on orthographic projections and the active sketch. CADENA is trained on a large-scale synthetically generated dataset rather than a limited collection of real manufacturing CAD models, enabling the use of a vision-language model with approximately thirty times more parameters. The training procedure combines supervised fine-tuning with online reinforcement learning, where rewards are computed from the geometry produced by executing the predicted program instead of offline sequence modeling. Furthermore, geometric discrepancies are encoded as a single image, making editing cues directly observable, rather than requiring comparison across multiple independent views. For the feedback, we use only the latest constructed shape, as opposed to the full history of shapes and operations. Finally, CADENA achieves substantially higher reconstruction accuracy than SOV-CAD across reported evaluation metrics (Appendix~\ref{sec:supp:sovcad}).

\paragraph{Benchmarks and metrics.} Reverse engineering is almost universally evaluated on the DeepCAD~\citep{deepcad} and Fusion 360~\citep{fusion360} test splits, both dominated by sketch--extrude parts. Corpora of real mechanical components exist --- MCB~\citep{mcb}, TMCAD~\citep{brt} and CADNet~\citep{cadnet} --- but they were assembled for classification, retrieval and machining-feature recognition, and consist largely of simple primitives and near-duplicates, so they are rarely used to evaluate reconstruction. Both metrics in use measure agreement of \emph{shape}, which is not what reverse engineering is for: IoU scores how far the prediction fills the target's volume and Chamfer distance how close the sampled points lie, and a reconstruction assembled from the wrong primitives satisfies either --- a cylinder approximated by an extruded polygon, or in the limit a shape packed with voxels, scores well while the program that produced it is wrong. IoU is in addition undefined unless both meshes are watertight, which excludes much of any real mechanical corpus, and both are averaged only over reconstructions that build, so a method that fails on the hard parts looks strong on the easy remainder. Recent work closes parts of this --- IterCAD~\citep{itercad} proposes a tolerance-recall curve free of survivor bias, HistCAD~\citep{histcad} measures whether edits preserve design intent, CAD-MLLM~\citep{cadmllm} adds topology-quality measures, and text-conditioned evaluation has moved toward executable tests~\citep{cadtests}, difficulty-stratified prompts~\citep{text2cadbench} and manufacturability rubrics~\citep{muse, cadjudge}. We address all three (Section~\ref{sec:bench}). CADENA-Bench distils the three mechanical corpora above into 3396 parts, discarding trivial and duplicate geometry and reporting per family rather than as one average. GMS matches points by normal as well as by position, so agreement requires the surfaces themselves to be of the same kind rather than merely to occupy the same space --- and a reconstruction whose surfaces are right is one whose construction tree is right. It is also defined for open geometry, so no part is dropped as unmeasurable. 

\section{Method}
\label{sec:method}

\begin{figure*}[t]
    \centering
    \includegraphics[width=\textwidth]{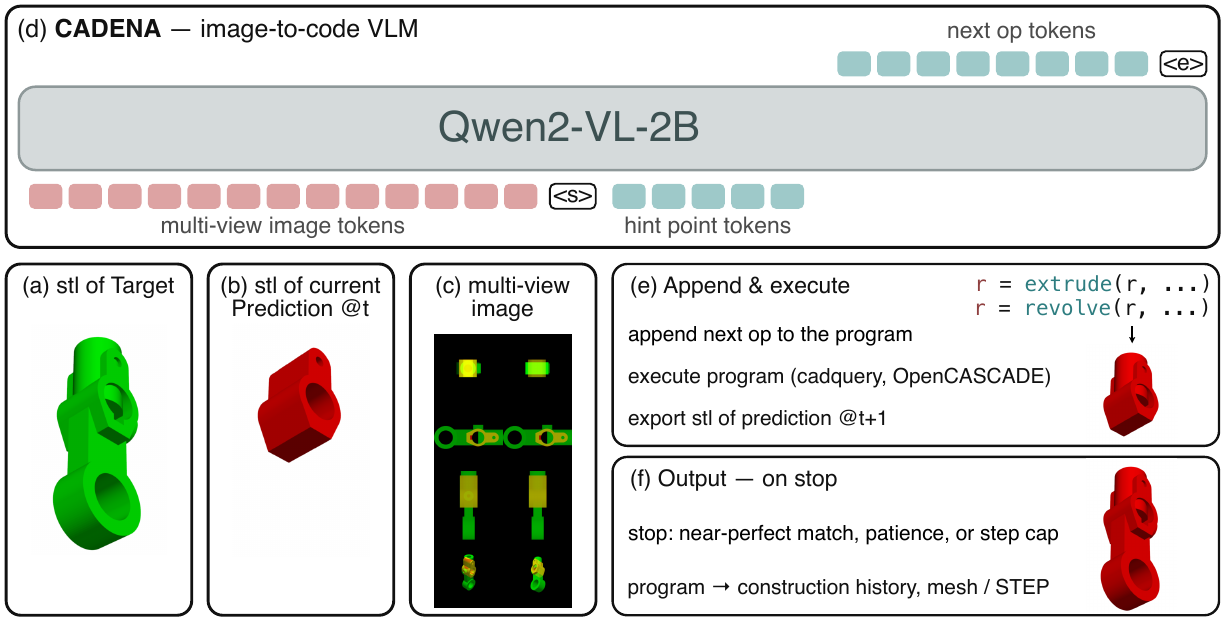}
    \caption{\textbf{CADENA overview.} The target mesh (green channel) and the current partial prediction (red channel) are rendered into a single aligned multi-view image consisting of six orthographic views (with depth encoded as color intensity) and two isometric views. Predominantly green regions indicate missing geometry, while predominantly red regions indicate excess material. Given this image and a hint point specifying the region to modify, the image-to-code VLM predicts the next operation, which is appended to the program. The updated program is then executed, producing a new partial build, and the process repeats until the stopping criteria are met. }
    \label{fig:method}
\end{figure*}

\subsection{Problem formulation}
Given a mesh $M$ of a mechanical part, we seek an editable parametric program $P$ whose execution reproduces $M$. We write $P$ as a sequence of operations $P = (o_1, \ldots, o_T)$ and define the state after $t$ steps as the prefix $P_t = (o_1, \ldots, o_t)$, whose execution yields a partial solid $S_t = \mathrm{exec}(P_t)$, with $S_0 = \varnothing$. Reconstruction is then a sequential decision problem: at each step the policy observes the target and what it has built so far, and appends one operation,
\begin{equation}
o_{t+1} \sim \pi\bigl(\cdot \mid \mathcal{R}(M, S_t)\bigr), \qquad P_{t+1} = P_t \oplus o_{t+1},
\end{equation}
where $\mathcal{R}$ is the observation function defined below. Crucially $\pi$ is not conditioned on the program text written so far: the observation is derived entirely from the target and the current build. The partial build $S_t$ is the sole carrier of history, which makes the policy a function of geometry rather than of code.

\subsection{Stepwise generation with visual feedback}
The observation $\mathcal{R}(M, S_t)$ is a single multi-view image that superimposes target and current build in separate color channels. Both meshes are rendered in a shared coordinate frame from the same eight viewpoints following the protocol of CADReasoner~\citep{cadreasoner} and CADEvolve~\citep{cadevolve}: the six axis-aligned directions ($\pm X$, $\pm Y$, $\pm Z$) using orthographic projections with depth encoded as color intensity, and two isometric views. The target is rendered as a green-channel overlay and the partial prediction as a red-channel overlay. The eight views are tiled into one $504 \times 1008$ image ($252 \times 252$ per view, a multiple of the vision encoder's patch size; rendered at twice that resolution and downsampled). This representation makes the difference between the two meshes directly readable: yellow indicates regions where the projections overlap (with the relative green/red intensity encoding depth mismatches), predominantly green regions mark missing geometry, and predominantly red regions indicate excess material (Fig.~\ref{fig:method} (c)). 

Alongside the rendered image, the policy receives a \emph{hint point}: a 3D coordinate provided as text. The hint disambiguates the next editing step when multiple disconnected discrepancy regions exist between the target mesh and the current prediction. Whenever possible, the hint is placed near the boundary between the existing and missing geometry, biasing the policy to extend the already constructed shape rather than begin a disconnected component.

During training, the ground-truth program and the intermediate shape after step $t+1$ are available. We uniformly sample points from the surface of this shape and compute the distance from each point to the mesh after step $t$. Points whose distance exceeds a threshold $\tau$ are considered to belong to the newly added geometry. The hint point is sampled uniformly from the subset of such points whose distances lie in the interval $(\tau, 3\tau)$. This favors points near the interface between the newly added and existing geometry. If this subset is empty, we instead sample a random point from the target mesh.

At inference time, the next intermediate shape is unknown, so the hint must be estimated from the target mesh and the current prediction. We first uniformly sample points from the target mesh and compute the distance from each point to the predicted mesh. Points farther than $\tau$ are treated as discrepancy candidates.
To identify spatially coherent discrepancy regions, we consider a sequence of progressively decreasing distance thresholds $\tau'$. For each threshold, we retain only candidate points whose distance to the predicted mesh exceeds $\tau'$, construct a 6-nearest-neighbor graph over the retained points, and extract its connected components. Large values of $\tau'$ isolate the cores of the most pronounced discrepancies and prevent distinct regions from being connected through points closer to the prediction. As $\tau'$ decreases, additional discrepancy regions become visible, although previously separate regions may begin to merge. We retain only newly discovered components and represent each component by the point within it that is farthest from the predicted mesh. Different components can be used to initialize different beam-search branches. For greedy decoding, we select the component whose representative is farthest from the prediction. Rather than using this representative directly as the hint, we move it toward the boundary between the reconstructed and missing geometry. Starting from the representative, we perform a greedy walk on a 6-nearest-neighbor graph constructed over all sampled target points. At each step, the walk moves to the neighboring point with the smallest distance to the predicted mesh. It terminates when it reaches a point whose distance is below $2\tau$, or after at most twice as many steps as there are points in the selected component. If the distance criterion is not reached, the visited point closest to the predicted mesh is returned. During the first iteration, when no prediction is yet available, the hint is sampled uniformly from the target mesh.

\begin{figure*}[t]
    \centering
    \includegraphics[width=\textwidth]{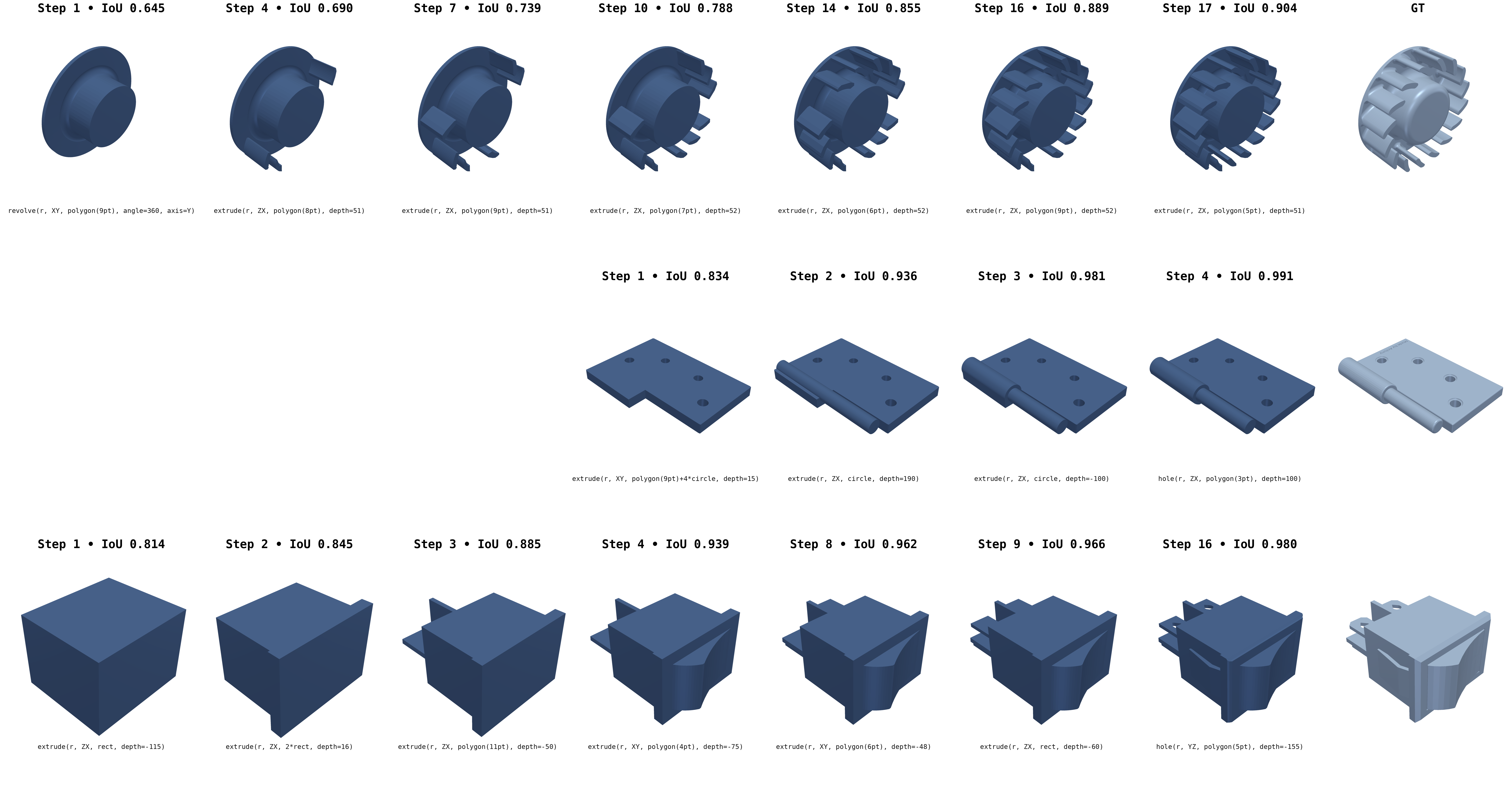}
    \caption{\textbf{Stepwise reconstruction by CADENA.} Each row follows one part from its first operation to the returned program. Columns are selected steps, annotated with the step index and the IoU of the build against the target at that point; the DSL line emitted at the step appears beneath each panel, and the rightmost column is the ground truth. The three parts are returned after 17, 4 and 16 operations. The model establishes bulk geometry first and adds detail later --- teeth, holes and pockets appear only once a body exists to cut them from --- and IoU rises along every row, since a prefix that lowers it is never the one selected. }
    \label{fig:chains}
\end{figure*}

An operation is a \emph{block} of CAD commands terminated by a three-dimensional operation --- extrude, revolve, sweep, loft, shell, chamfer, fillet, or gear --- so that every step ends in a state that can be executed and rendered; the vocabulary as actually emitted is listed in Table~\ref{tab:dsl}. In practice such a block is written as a single DSL line, and we describe it as one line throughout. Figure~\ref{fig:chains} shows complete reconstruction chains, with the observation and the emitted operation at each step.

\subsection{DSL and execution}
Operations are single lines of a CadQuery-based DSL (full grammar and per-operation argument conventions in Appendix~\ref{sec:supp:dsl}), executed with OpenCASCADE. The DSL exists to make a step predictable \emph{in isolation}. A step must be emitted from the two images alone, without reading the code that produced the current state, so the syntax cannot refer to variables bound by earlier lines. We therefore let the construction tree grow through a single variable $r$: every operation consumes $r$ and rebinds it,
\begin{center}
\texttt{r = extrude(r, point, plane, "sketch(\ldots)", h)},
\end{center}
so that any step is a well-formed continuation of any prefix. This is what makes the policy's observation sufficient: with no free variable names to resolve, the next line depends on the current geometry and not on the particular text that produced it. Two constructs are exceptions and are stated for completeness: edge operations are written as selector chains, \texttt{r = r.edges(\ldots).fillet(radius)}, which still consume and rebind $r$; and \texttt{gear} consumes an auxiliary workplane bound on the preceding line rather than $r$, so it occurs only as a first operation. Every other operation follows the single-variable form.

Coordinates in a DSL line are absolute and integral, because generator, policy and evaluation all work in one normalised frame. Every final geometry and its CAD model are rescaled so that the largest bounding-box dimension equals a fixed world size and coordinates are rounded to integers, so values lie in $[-100, 100]$. Predictions are scored in this same frame rather than being re-centred and re-scaled per part, which is what makes absolute scale and placement errors count in the numbers reported in Section~\ref{sec:results}. 

\subsection{Training}
\paragraph{Supervised fine-tuning.} Supervision comes from a rule-based procedural generator that samples valid CAD programs together with the meshes they produce, covering extrude, revolve, loft, sweep, shell, hole, gear, spring, and edge operations. Because the generator owns the construction history, it can emit \emph{stepwise} supervision directly: a program is cut at a checkpoint, the prefix is frozen and executed to give the current state, and the continuation supplies the target operation. Each training sample is therefore a single step --- the target shape, the partial build produced by the prefix, and the code of the next operation --- in exactly the form the policy consumes at inference.

Training proceeds in two stepwise stages. The warm-up stage uses 1.86 million samples drawn from short programs of at most two operations, teaching the model the DSL and the correspondence between residual geometry and operation parameters on problems where little history has accumulated. The main stage uses 18.0 million samples from programs containing up to twelve operations, pooled across generator runs of increasing length (2, 4, 6, 8, 10, and 12 operations). This exposes the model to partial constructions at every depth rather than only near the beginning of a program. The main-stage model is initialized from the weights learned during the warm-up stage. The policy is a Qwen2-VL vision--language model, and is trained using teacher-forced next-operation prediction for $2$ epochs per stage, with a batch size of $64$, an initial learning rate of $1.2 \times 10^{-4}$, and a cosine learning-rate schedule.

\paragraph{Reinforcement learning: reward.}
We then fine-tune the stepwise policy with RL against the program environment. The
policy generates trajectories step by step, executing each operation in the loop,
but because execution dominates the cost of a rollout we do not optimise whole
trajectories: training uses one-step signals, optimising the choice of the next
operation relative to a reference trajectory that the policy generates itself. The
reward is the volumetric IoU between the solid built so far and the target mesh,
computed directly on the meshes; a candidate whose program fails to execute
receives zero. Since every intermediate state is itself a solid, this reward is
available after each step, and the supervision is purely geometric: it requires the
target mesh and nothing else, in particular no ground-truth program. Rewarding
executed geometry rather than token overlap also makes validity part of the
objective, so validity is learned rather than enforced by post-hoc filtering.
\begin{figure*}[!t]
    \centering
    \includegraphics[width=0.85\textwidth]{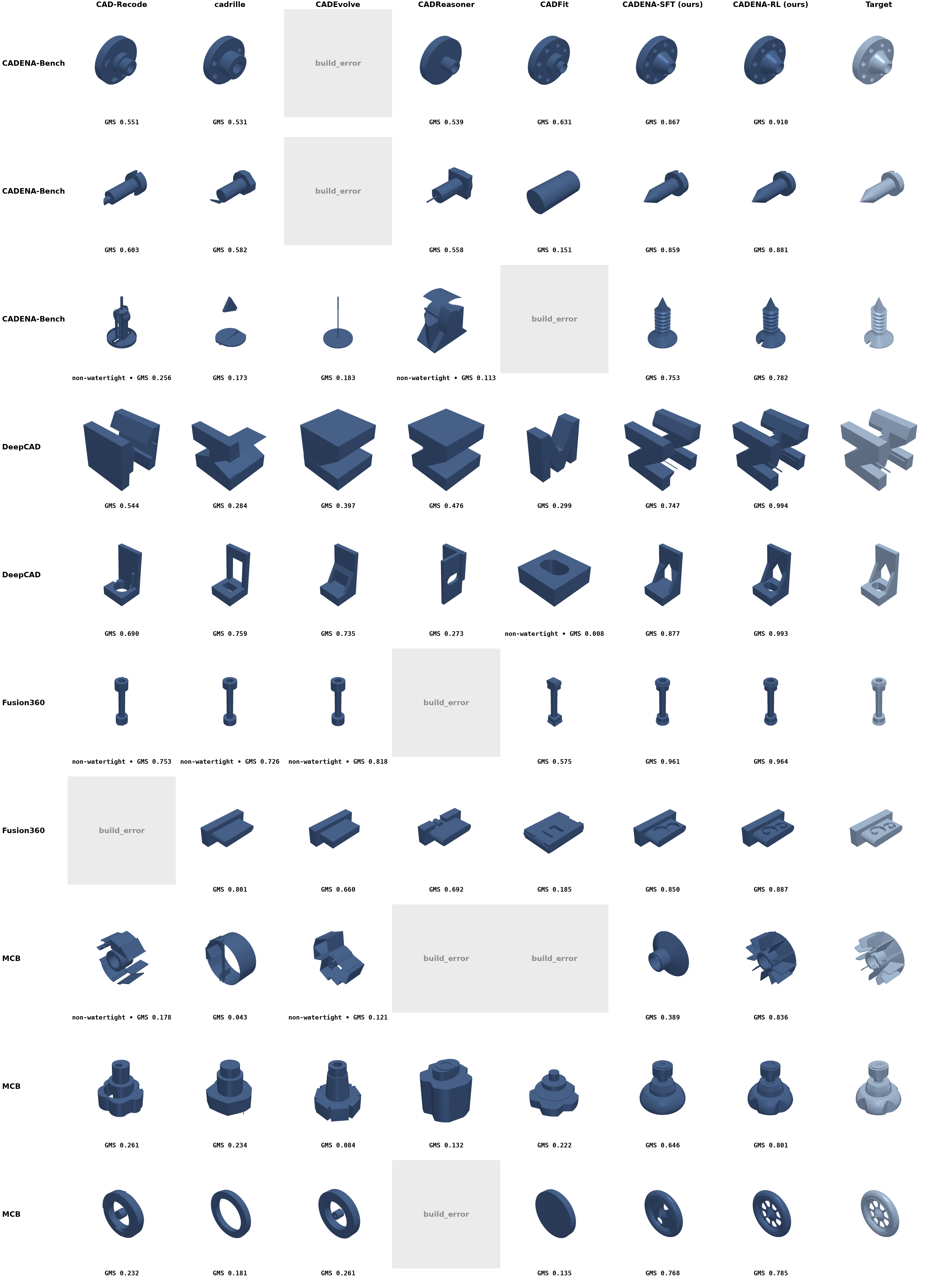}
    \caption{\textbf{Qualitative comparison across methods.} Each row is one input part, labelled with the dataset it comes from; each column is one method, with the target in the rightmost column. The GMS of that single reconstruction is printed beneath each panel. Grey panels mark predictions that failed to build, and predictions that build without being watertight are annotated as such; both count toward the invalid rate and are excluded from the means in Table~\ref{tab:external}. The failures are not spread evenly --- they concentrate on the turned and patterned parts of MCB and CADENA-Bench, which is the same pattern the per-family results show. }
    \label{fig:qualitative}
\end{figure*}
\paragraph{Rollout construction.}
RL uses ${\sim}4$k meshes taken from \emph{training} splits only, so no test
geometry is seen during training: 1{,}500 from MCB, sampled uniformly at random
across its object classes, and 2{,}500 sampled at random from DeepCAD and
Fusion360. MCB provides no reference programs at all, which the geometric reward
makes irrelevant here. At each iteration we take a batch of 16 target meshes.
For each mesh the current policy first produces a reference trajectory by greedy
decoding, capped at 10 operations, though most trajectories in a batch terminate
well before the cap is reached. Every prefix state along this trajectory is then
branched into 32 candidate continuations sampled at temperature $T{=}1.0$; each
candidate is executed and scored, forming one group per step. A group is discarded
if no candidate strictly improves IoU over the prefix state it was branched from,
and likewise if the standard deviation of its rewards falls below $10^{-4}$, since
such a group carries no learning signal.

\paragraph{Group selection and updates.}
For the surviving groups, advantages follow the standard GRPO normalisation:
rewards are centred by the group mean and scaled by the group standard deviation.
From each group we keep the two candidates with the largest absolute advantage, so
that both clearly better and clearly worse continuations contribute to the update.
Groups are processed in decreasing order of their pre-normalisation reward standard
deviation, and the resulting list of selected candidates is truncated to the
largest multiple of the effective batch size ($3$ devices $\times\,4$ per device
$\times\,4$ gradient-accumulation steps $=48$); this ordering ensures that what the
truncation drops is the least diverse, and hence least informative, part of the
batch. The loss includes no KL penalty against the reference policy
($\beta{=}0$): a KL term is ahead on most metrics on the 1000-example subsamples,
but the advantage is not confirmed on the full test splits, where the difference
falls within noise (App.~\ref{app:rl-experiments}). Training runs for 20 epochs
over the ${\sim}4$k meshes, i.e.\ $5{,}000$ batches. The number of updates per
batch is not fixed: the greedy trajectory sets how many branching points a mesh
contributes, and its length is decided by decoding rather than set in advance, so
the number of selected candidates --- and hence of effective batches of $48$ ---
varies. In practice this comes out at $3$--$5$ updates per batch, or roughly
$20$k policy-gradient updates in total.

\subsection{Inference}
Decoding is greedy, one operation per step, with a budget of 20 operations.
Despite training only on 10-step trajectories, in our experiments the policy
maintains its performance when decoded to this longer budget, extrapolating to a
horizon twice as long as the one it was trained on. The longer budget is
affordable at inference for the reason a shorter one is necessary during RL:
there, each step branches into 32 candidate continuations, so a shorter horizon
keeps rollout cost tractable, whereas a single greedy pass is cheap enough to
afford a longer budget for more complex shapes.

Because every prefix is executable, the loop yields not one candidate but a chain
of them: the built solid after each step is scored against the target, and we
return the prefix that maximises IoU. Selection therefore never consults a label
--- the target mesh is the \emph{input} to reverse engineering, so comparing the
current build against it is available at test time. This also means the method
degrades gracefully: a late operation that damages the reconstruction is simply not
selected, so additional steps cannot make the returned program worse. Selecting the best prefix rather than the last is worth little on its own (Appendix~\ref{sec:supp:ablations}); the gain comes from conditioning each operation on the residual, not from the selection rule.

Our main results use greedy decoding, so that every method is compared under the
same single-sample protocol (Section~\ref{sec:results}). Sampling at $T{=}1.0$ is
the stronger rule, and we report it alongside
(Section~\ref{sec:decoding}).

\section{CADENA-Bench and the GMS metric}
\label{sec:bench}

\subsection{CADENA-Bench}
CADENA-Bench is 3396 unique mechanical parts drawn from MCB~\citep{mcb}, TMCAD~\citep{brt}, and CADNet~\citep{cadnet}, deduplicated and classified into six part families. It is released at \url{https://huggingface.co/datasets/kulibinai/cadena-bench}.

\paragraph{Sources and deduplication.} The initial pool contained approximately 60k parts and a substantial number of duplicates, both within individual sources and across them. To remove them we encoded each mesh with a transformer encoder trained to predict MCB categories, clustered the resulting embeddings with DBSCAN, and manually inspected all ${\sim}300$ resulting clusters, retaining one representative per group. We counted as duplicates not only exact geometric copies but also mirrored variants and parts differing only by minor parametric modifications, since none of these adds geometric diversity to the benchmark. The benchmark targets individual manufactured parts, so objects representing assemblies of several joined components were excluded. Parts too simple to exercise a reconstruction method --- bare primitives and near-primitives --- were discarded on the same grounds. 

\paragraph{Part families.} Each surviving part carries one of six family labels --- shafts \& bushings, gears \& bearings, housings \& frames, flat \& levers, springs \& fasteners, tooling \& gauges --- coarse classes corresponding to the standard engineering parts classification. The taxonomy is deliberately coarse so that it can be applied consistently across three heterogeneous sources. Labels were predicted by the Qwen3.5-122B-A10B vision--language model from two isometric views of each part (six orthographic projections and two isometric), with the model instructed to classify using only the geometry visible in the renders. The prompt first asks whether the object is a single manufactured part, an assembly, or unclassifiable geometry; for individual parts it then asks for exactly one of the six families, judged primarily by the likely functional purpose of the part, with the orthographic views used to assess proportions, silhouette, and features indicative of a body of revolution.

\paragraph{What the labels are, and are not.} The labels characterise the composition of the benchmark and organise the reporting; they enter neither the method, nor GMS, nor any number reported in \S\ref{sec:results}. What they do bound is how finely the per-family columns can be read, so we measured their agreement with human annotation: on a random subset of 254 manually annotated parts the model labels agreed with the manual labels in 70\% of cases. Approximately two thirds of the disagreements fall between three pairs of neighbouring families whose boundaries in the original classification depend on the functional purpose of the part and not on geometry alone --- bodies of revolution vs.\ fasteners, housings vs.\ planar parts, and parts with vs.\ without engagement elements. A per-family number is therefore a statement about a family as labelled here, not about a certified partition of mechanical parts.

\subsection{The GMS metric}
\label{sec:gms}
Section~\ref{sec:related} argued that agreement of shape does not establish that the construction tree was recovered. The \emph{Generalized Match Score} (GMS) makes surface type the observable instead: it matches points by normal as well as by position, so a cylinder is matched only by a cylinder and not by a prism filling the same volume. We take agreement of surface type as evidence that the tree is correct --- an assumption, but the weakest we know of that separates a recovered program from a recovered silhouette. GMS is defined for open geometry, so no part is dropped as unmeasurable.

Let $A$ and $B$ be $N$ points sampled on the reference and reconstructed surfaces, each carrying a unit normal. Given a relative distance tolerance $\tau$ and an angular tolerance $\alpha$, a point $p \in A$ is matched if some $q \in B$ satisfies both $\lVert p - q \rVert \le \tau$ and $n_p \cdot n_q \ge \cos\alpha$; the matched subset is written $A \cap_{\tau,\alpha} B$, and symmetrically for $B$. Recall and precision are the matched fractions,
\begin{equation}
\mathrm{Rec}(\tau,\alpha) = \frac{\lvert A \cap_{\tau,\alpha} B \rvert}{\lvert A \rvert}, \qquad
\mathrm{Prec}(\tau,\alpha) = \frac{\lvert B \cap_{\tau,\alpha} A \rvert}{\lvert B \rvert},
\end{equation}
capturing missing and excess geometry respectively, and are combined harmonically into a single agreement score at that pair of tolerances,
\begin{equation}
g(\tau,\alpha) = 2\bigl(\mathrm{Rec}(\tau,\alpha)^{-1} + \mathrm{Prec}(\tau,\alpha)^{-1}\bigr)^{-1} \in [0,1].
\end{equation}
The harmonic form means a reconstruction cannot score well by covering the target while adding spurious material, or by reproducing a small fragment precisely.

No single pair $(\tau,\alpha)$ yields values that are comparable across parts of differing geometric complexity, and in our sensitivity analysis it is the angular tolerance that dominates the score. We therefore fix the distance tolerance, integrate over the angular one, and report the \emph{normalised area under the tolerance curve},
\begin{equation}
\mathrm{GMS} = \frac{100}{\alpha_{\max}} \int_{0}^{\alpha_{\max}} g(\tau, \alpha) \, \mathrm{d}\alpha,
\qquad \alpha_{\max} = 25^\circ,
\end{equation}
evaluated at fixed $\tau = 0.05$ and $N = 8192$ points, with the integral approximated on a $0.2^\circ$ grid. Both meshes are normalised into $[0,1]^3$ before sampling, which fixes $\tau$ by the sampling density rather than leaving it to be chosen: partitioning that cube into $N$ equal voxels gives an edge of $N^{-1/3} \approx 0.05$, the mean spacing of the sampled points. A tighter tolerance would measure sampling noise, and a looser one would blur geometry the sampling can still resolve. Since $g \in [0,1]$, the score lies in $[0, 100]$, and larger is better: a faithful reconstruction reaches high $g$ at tight tolerances and fills the area under the curve, whereas a poor one only matches once the tolerance is loose. 

Because the score depends only on relative distances and normal agreement, it is scale-normalised, tolerant to sampling density, and --- unlike IoU --- defined for open, non-watertight geometry, which is what makes it usable on every part of CADENA-Bench rather than on the watertight subset.

GMS serves for evaluation only. We tested it directly as an RL reward, and the outcome was self-defeating: the resulting policy improved on GMS while leaving Chamfer distance and IoU no better, and markedly worse on the MCB subsample (Appendix~\ref{app:gms}). Every reported policy is therefore trained against volumetric IoU and selects its returned prefix by the same criterion, with GMS entering only at evaluation.

\section{Experiments}
\label{sec:results}

\subsection{Experimental setup}
We evaluate on the DeepCAD, Fusion360, and MCB test meshes and on CADENA-Bench (\S\ref{sec:bench}). All methods run in greedy single-sample mode from their released checkpoints.

A prediction is valid if it builds into a watertight solid; invalid predictions count toward the invalid rate (IR) and are excluded from all means. Table~\ref{tab:benchmark} reports GMS per part family; the remaining metrics for CADENA-Bench, including IR, are broken out in full in Appendix~\ref{sec:supp:fulleval}.

\paragraph{Baseline configurations.} Each baseline is run from its released checkpoint in the input modality its authors report as strongest: cadrille from images, CAD-Recode from point clouds, and CADReasoner with five refinement rounds. CADEvolve is evaluated at RL1, the only checkpoint publicly released; their RL2 variant adds the MCB training split and is not available. BenchCAD is the exception: there we evaluate only CADENA, scoring it with the procedure its authors specify, and take every other figure as published --- from the BenchCAD paper and from the leaderboard on their website. We ran none of those systems ourselves. 

\paragraph{Two departures from the standard protocol.} Our evaluation differs from the one used in prior CAD reverse-engineering papers in two ways. Both are deliberate, and both make the task harder rather than easier to score well on.

First, predictions are evaluated in the fixed frame in which the models operate, rather than being re-centred and re-scaled by their own bounding box. A program that reproduces the shape of a part but places it elsewhere, or builds it at another size, has not reconstructed the input; per-prediction re-normalisation removes exactly that error before the metric can see it. Under our protocol absolute scale and placement errors count. CADFit is the single exception --- it fits the input mesh directly and so cannot commit such an error by construction, and is scored with prediction and target normalised alike (note to Table~\ref{tab:external}).

Second, each metric is averaged only over the set on which it is defined: IoU over parts with watertight ground truth, every metric over valid predictions. Averaging instead over all parts requires substituting some value wherever the metric has none, so the published mean mixes measured reconstruction quality with the substitution rule, and the reader cannot separate them. We prefer to average over the set where the quantity exists and to report separately, as IR, how large the excluded set is.

\paragraph{What the two changes are worth.} They act in opposite directions and are of very different size. Writing each chain as \emph{published protocol} $\to$ \emph{well-defined averaging set} $\to$ \emph{fixed frame}, cadrille goes $47.6 \to 67.0 \to 66.2$ IoU on MCB and $92.2 \to 92.9 \to 89.7$ on DeepCAD. The first step of the MCB chain is large enough that it should not pass as a parenthetical: restricting the average to the parts where IoU is defined moves a published baseline number from 47.6 to 67.0. This is not a correction of cadrille --- its own protocol reproduces its own published value exactly, as below --- but a change of averaging convention, and it is the clearest evidence we have that on real mechanical parts the protocol, and not only the method, decides much of what gets reported. Neither convention scores an undefined IoU as zero; both drop the part. What differs is which parts remain: we average only over those whose target and prediction are both non-degenerate solids, and parts that survive that test are easier on average than the full set, which is what moves the number.

\subsection{Main results}
Table~\ref{tab:external} reports the external benchmarks, Table~\ref{tab:benchmark} the per-family results on CADENA-Bench, and Table~\ref{tab:benchcad} the BenchCAD leaderboard. Figure~\ref{fig:qualitative} contrasts reconstructions of the same parts across methods, and Appendix~\ref{sec:supp:failures} analyses the failure modes they show.

\paragraph{External datasets are saturated.} On DeepCAD every method lands within a few points of every other: CD between 0.15 and 0.18, IoU between 89.7 and 96.1, GMS between 89.8 and 97.0. The ranking is decided by differences comparable to the noise of the evaluation itself, so a method that reconstructs mechanical parts poorly is nearly indistinguishable from one that does it well.

\paragraph{Real mechanical parts are dramatically harder.} Moving from DeepCAD to CADENA-Bench costs every learned method roughly half its GMS: cadrille $94.8 \to 49.8$, CAD-Recode $92.9 \to 48.9$, CADEvolve $95.3 \to 52.9$, CADReasoner $94.9 \to 52.9$. CADENA-RL falls from $97.0$ to $67.0$, so its margin over the strongest baseline widens from $1.7$ points on DeepCAD to $12.2$ on CADENA-Bench. Test sets built from sketch--extrude corpora understate the difficulty of the task they are taken to measure.

\paragraph{The degradation is uneven across part families.} Gears \& bearings and springs \& fasteners are the weakest families for every learned method, both dominated by revolved and patterned features rather than prismatic ones. CADENA-RL leads five of the six; the exception is gears \& bearings, where CADFit's direct fitting reaches $61.3$ against our $58.1$. Its largest margins over the learned baselines fall on springs \& fasteners ($63.4$ against $47.9$) and shafts \& bushings ($73.0$ against $59.6$). Tooling \& gauges holds 17 parts, and we do not read that column as a ranking.

\paragraph{Validity separates the methods more than accuracy does.} Invalid rates span two orders of magnitude on the same inputs: cadrille almost always returns a buildable watertight solid ($0.3$--$1.5\%$), whereas CADReasoner fails on $31.7\%$ of MCB parts and CAD-Recode on $21.5\%$. About half of CADReasoner's failures are degenerate three-point arcs that OpenCASCADE refuses to build, and a sixth are programs truncated mid-expression by repetition loops. Because invalid parts are excluded from every mean, a method can post competitive CD or IoU while reconstructing only the subset it finds easy; per-family invalid rates for CADENA-Bench are given in Appendix~\ref{sec:supp:fulleval}. CADENA-RL combines the accuracy lead with an invalid rate of at most $1.2\%$ on the three external datasets, which we attribute to the geometric reward: a candidate that does not build scores zero, so validity is optimised rather than filtered for afterwards.

\paragraph{Direct optimisation inverts the profile.} CADFit is the one baseline that does not learn a generator, and its profile is inverted. It is the weakest method on DeepCAD ($89.8$ GMS) yet the strongest baseline on MCB ($60.1$) and CADENA-Bench ($54.8$), because fitting geometry directly does not depend on the training distribution matching the test parts. Its IoU is the highest of any baseline on Fusion360 and MCB ($88.2$ and $75.8$), though CADEvolve's $92.4$ is higher on DeepCAD. The cost is validity: under a fixed compute budget it leaves $25.6$--$29.7\%$ of the external parts without a valid reconstruction, and since the parts that exhaust the budget are the slowest and fitting time grows with complexity, its accuracy is measured over an easier-than-average subset. That invalid rate is also a different object from the others in the table: it records an unconverged search rather than a malformed program, and cannot be diagnosed by inspecting the output. This is the mirror image of the learned methods' failure mode --- they degrade smoothly on hard parts, whereas optimisation either reproduces a part closely or returns nothing.

Several recent systems target the same setting but could not be included in the tables above; Appendix~\ref{sec:supp:omissions} states why in each case.

\begin{table}[t]
\centering
\caption{\textbf{CAD reverse engineering on external benchmarks.}
Median CD$\downarrow$ ($\times 10^{3}$) at 8k and 30k sampled points, mean IoU$\uparrow$ (\%), GMS$\uparrow$ (\%), and IR$\downarrow$ (invalid rate, \%) on DeepCAD, Fusion360, and MCB.
A prediction is invalid if it fails to build or is not watertight; invalid predictions are counted in IR and excluded from all means. All metrics are evaluated in the fixed normalized frame used for all methods; IoU is additionally restricted to parts with watertight ground truth.}
\label{tab:external}
\small
\setlength{\tabcolsep}{2.5pt}
\resizebox{\linewidth}{!}{
\begin{tabular}{lccccc ccccc ccccc}
\toprule
& \multicolumn{5}{c}{DeepCAD} & \multicolumn{5}{c}{Fusion360} & \multicolumn{5}{c}{MCB} \\
\cmidrule(lr){2-6}\cmidrule(lr){7-11}\cmidrule(lr){12-16}
Method & CD$_{8k}\downarrow$ & CD$_{30k}\downarrow$ & IoU$\uparrow$ & GMS$\uparrow$ & IR$\downarrow$
       & CD$_{8k}\downarrow$ & CD$_{30k}\downarrow$ & IoU$\uparrow$ & GMS$\uparrow$ & IR$\downarrow$
       & CD$_{8k}\downarrow$ & CD$_{30k}\downarrow$ & IoU$\uparrow$ & GMS$\uparrow$ & IR$\downarrow$ \\
\midrule
CAD-Recode   & 0.17 & 0.055 & 91.4 & 92.9 & 7.9
             & 0.17 & 0.060 & 87.2 & 85.8 & 13.6
             & 0.66 & 0.524 & 71.1 & 54.7 & 21.5 \\
cadrille     & 0.17 & 0.059 & 89.7 & 94.8 & \textbf{0.3}
             & 0.17 & 0.066 & 84.8 & 86.8 & \textbf{0.8}
             & 0.85 & 0.720 & 66.2 & 55.0 & \underline{1.5} \\
CADReasoner  & \underline{0.16} & 0.049 & 91.0 & 94.9 & 3.0
             & 0.15 & 0.050 & 86.3 & 88.6 & 7.2
             & 1.44 & 1.290 & 69.0 & 55.2 & 31.7 \\
CADEvolve    & \underline{0.16} & 0.050 & \underline{92.4} & \underline{95.3} & 0.9
             & 0.16 & 0.056 & 88.1 & 88.2 & 2.7
             & 0.63 & 0.500 & 72.9 & 58.3 & 9.2 \\
CADFit$^\dagger$ & 0.18 & 0.052 & 91.4 & 89.8 & 27.5
             & 0.17 & 0.054 & 88.2 & 84.6 & 25.6
             & 0.48 & 0.365 & \underline{75.8} & 60.1 & 29.7 \\
\midrule\midrule
CADENA-SFT   & \underline{0.16} & \underline{0.048} & 91.7 & \underline{95.3} & 2.6
             & \underline{0.14} & \underline{0.046} & \underline{88.8} & \underline{90.8} & 3.8
             & \underline{0.43} & \underline{0.259} & 75.2 & \underline{63.4} & 12.0 \\
CADENA-RL    & \textbf{0.15} & \textbf{0.042} & \textbf{96.1} & \textbf{97.0} & \textbf{0.3}
             & \textbf{0.12} & \textbf{0.038} & \textbf{94.1} & \textbf{93.3} & \underline{1.2}
             & \textbf{0.22} & \textbf{0.093} & \textbf{88.2} & \textbf{73.7} & \textbf{0.7} \\
\bottomrule
\end{tabular}}
\tabnote{$^\dagger$CADFit fits the input mesh directly rather than generating a program in a learned coordinate convention, so it cannot commit a scale or placement error by construction; it is therefore scored with prediction and target normalised alike. CADFit is an optimisation loop rather than a feed-forward model, and we ran it under a fixed compute budget; parts whose fit had not converged when the budget expired are counted in IR (DeepCAD 1635/8046, Fusion360 320/1725, MCB 1149/5000). Because the unconverged parts are the slowest ones, and fitting time grows with geometric complexity, its means are computed over an easier-than-average subset and should be read as optimistic.}
\end{table}

\begin{table}[t]
\centering
\caption{\textbf{Per-category GMS$\uparrow$ on CADENA-Bench} (3396 mechanical parts, six part families). GMS is averaged over valid predictions (buildable and watertight), as in Table~\ref{tab:external}.}
\label{tab:benchmark}
\small
\setlength{\tabcolsep}{4pt}
\begin{tabular}{lccccccc}
\toprule
& Shafts \& & Gears \& & Housings \& & Flat \& & Springs \& & Tooling \& & \\
Method & bushings & bearings & frames & levers & fasteners & gauges & All \\
& \scriptsize(762) & \scriptsize(749) & \scriptsize(359) & \scriptsize(464) & \scriptsize(1046) & \scriptsize(17) & \scriptsize(3396) \\
\midrule
CAD-Recode   & 53.8 & 39.1 & 56.1 & 58.8 & 45.3 & 38.7 & 48.9 \\
CADReasoner  & 55.6 & 46.9 & 56.4 & \underline{69.7} & 43.4 & 38.5 & 52.9 \\
cadrille     & 57.6 & 37.2 & 54.5 & 62.1 & 46.3 & 41.8 & 49.8 \\
CADEvolve    & 59.6 & 41.3 & \underline{59.4} & 66.4 & 47.9 & \underline{44.7} & 52.9 \\
CADFit$^\dagger$ & 60.8 & \textbf{61.3} & 48.5 & 66.2 & 42.3 & 23.1 & 54.8 \\
\midrule\midrule
CADENA-SFT   & \underline{70.8} & 47.8 & 58.1 & 66.8 & \underline{51.9} & 43.4 & \underline{57.8} \\
CADENA-RL    & \textbf{73.0} & \underline{58.1} & \textbf{70.7} & \textbf{76.8} & \textbf{63.4} & \textbf{60.0} & \textbf{67.0} \\
\bottomrule
\end{tabular}
\tabnote{$^\dagger$CADFit is scored with prediction and target normalised alike (see Table~\ref{tab:external}); its 638/3396 unconverged or invalid parts are excluded from these means (IR 18.8\%).}
\end{table}

\paragraph{Comparison against frontier models on BenchCAD.} Recent frontier model reports evaluate on BenchCAD~\citep{benchcad}: the Claude Opus~5 and GPT-5.6 system cards both publish Vision2Code scores, and the leaderboard additionally carries entries for Gemini~3.1~Pro. It is therefore the one setting in which a specialist reconstruction system can be placed beside general-purpose models under a protocol we did not define. Vision2Code scores a generated program by voxel IoU against the ground-truth solid and reports the fraction of programs that execute. We evaluate CADENA on the full released corpus ($17{,}895$ parts) and report it beside the published board in Table~\ref{tab:benchcad}; every other figure is taken as published. We restrict the comparison to voxel IoU and invalid rate, since BenchCAD's Chamfer distance and composite score depend on conventions it does not specify.

The comparison is not like-for-like, and the asymmetry favours us: entries on the board receive a single composite image, whereas CADENA consumes the target mesh and selects its returned prefix by IoU against that same mesh. The table should be read as a pattern across systems rather than as a ranking.

\begin{table}[t]
\centering
\caption{\textbf{Vision2Code on BenchCAD.} Voxel IoU$\uparrow$ and invalid rate$\downarrow$ (\%). The comparison is \emph{not} like-for-like and should not be read as a ranking; see the note below.}
\label{tab:benchcad}
\small
\setlength{\tabcolsep}{5pt}
\begin{tabular}{llccc}
\toprule
Model & Input & Graded by & Voxel IoU$\uparrow$ & IR$\downarrow$ \\
\midrule
\multicolumn{5}{l}{\emph{Frontier vision--language models}} \\
GPT-5.6 Sol (thinking)     & image & vendor   & 0.706 & --- \\
GPT-5.6 Luna (thinking)    & image & vendor   & 0.631 & --- \\
GPT-5.6 Terra (thinking)   & image & vendor   & 0.623 & --- \\
GPT-5.5 (thinking)         & image & vendor   & 0.444 & --- \\
Claude Mythos 5 (thinking) & image & vendor   & 0.384 & --- \\
Claude Opus 5 (thinking)   & image & vendor   & 0.366 & --- \\
Gemini 3.1 Pro (thinking)  & image & BenchCAD & 0.355 & 18.5 \\
Claude Opus 4.7 (thinking) & image & BenchCAD & 0.279 & 3.5 \\
\midrule
\multicolumn{5}{l}{\emph{Specialist CAD models}} \\
qwen3-2b-rl-iid            & image & BenchCAD & 0.752 & 1.1 \\
CADEvolve                  & image & BenchCAD & 0.750 & 7.3 \\

\midrule\midrule
CADENA-RL (ours)           & mesh  & ours     & \textbf{0.910} & \textbf{0.9} \\
\bottomrule
\end{tabular}
\tabnote{\textbf{This table compares systems that do not receive the same input, and the difference favours us.} BenchCAD's Vision2Code supplies each model with a single composite image of four orthographic views. CADENA consumes the target \emph{mesh}: it renders its own eight views from that mesh, and --- more consequentially --- selects the returned prefix by IoU against it. Since the mesh is also the ground truth BenchCAD scores against, our selection rule optimises the reported quantity directly, which no image-conditioned entry can do. The gap should therefore be attributed to the richer input and to test-time selection, not to the generator alone. \emph{Grading.} Rows marked \emph{BenchCAD} were re-graded by the benchmark authors from raw outputs; they publish \texttt{IoU-score} $=$ voxel IoU $\times$ \texttt{exec\%}, which we divide by \texttt{exec\%} to recover plain voxel IoU (this reproduces their blind-baseline row exactly). Rows marked \emph{vendor} are self-reported, published as voxel IoU only and not re-graded, so no execution rate is available and the value has not been independently verified. We list them because omitting them would understate the frontier: the strongest self-reported entry is closer to our result than any re-graded frontier entry. \emph{Sources.} The frontier rows are taken from the BenchCAD leaderboard, which is scored on a $1{,}000$-part subset; the specialist rows are taken from the BenchCAD paper and use the full corpus, as does ours. \emph{Tools.} The self-reported rows are tool-free configurations. The vendors' own reports give higher figures when the model is allowed tools and additional test-time compute --- $0.821$ for Claude Opus~5 and $0.678$ for Claude Mythos~5 --- which the leaderboard does not carry and which are not comparable with a single-pass system. \emph{Our row.} We grade ourselves with BenchCAD's published scorer, verified by reproducing their blind-image control exactly; their submission process accepts raw outputs for official re-grading, which we have not yet used. }
\end{table}

CADENA-RL reaches $0.910$ voxel IoU at $0.9\%$ invalid. The figure is not comparable with the mesh IoU we report elsewhere: BenchCAD voxelises both solids on a $64^3$ grid, which is coarser than our fixed-frame comparison and correspondingly more forgiving. The strongest reported frontier entry, GPT-5.6 Sol, reaches $0.706$; the frontier entries the benchmark authors re-graded from raw outputs top out at $0.355$, and the two specialist systems at $0.750$ and $0.752$, the latter being \texttt{qwen3-2b-rl-iid}, the benchmark authors' own baseline: a 2B vision--language model fine-tuned with reinforcement learning on data drawn from the benchmark's own distribution. The pattern across the board is that recovering industrial geometry, not emitting valid CadQuery, is what separates systems: re-graded frontier entries reach $96.5\%$ execution while their voxel IoU stays below $0.36$.

\subsection{Decoding: greedy versus sampling}
\label{sec:decoding}

Every other result we report uses greedy decoding, one sample per part. That is a
choice about comparability rather than about quality: the baselines are run
single-sample, so a sampled CADENA would be compared against them at an unequal
budget. Sampling is in fact the better decoding rule, and we give the difference
here rather than leave it implicit. Table~\ref{tab:decoding} reports the same two checkpoints decoded by sampling instead: at each operation we draw $E{=}12$ candidate continuations at $T{=}1.0$, execute them, and keep the one whose build scores highest against the target by IoU, exactly the criterion used to select the returned prefix. Sampling therefore costs twelve executions per operation where greedy costs one, which is the second reason the reported numbers are greedy: it is not a decoding rule the baselines were given either.

Sampling is no worse than greedy in any cell of the table, for either model on
any of the three datasets, and the difference is largest where greedy is weakest:
the SFT model on MCB moves from $63.3$ to $70.0$ GMS and from $12.04\%$ to
$0.36\%$ invalid, whereas the RL model on DeepCAD --- the strongest combination
in the table under either rule --- moves from $97.0$ to $97.4$. The ordering of
the two models is unchanged: RL is ahead of SFT under either decoding rule.

\begin{table*}[t]
\centering
\caption{Comparison of the SFT and RL models under greedy decoding and sampling
($T{=}1.0$, $E{=}12$). IoU and GMS are means reported in percent, CD values are
medians (computed with $8192$ and $30000$ sampled points), and IR is the fraction of
invalid generations. Best values per dataset are in bold.}
\label{tab:decoding}
\setlength{\tabcolsep}{3.5pt}
\renewcommand{\arraystretch}{1.1}
\resizebox{\textwidth}{!}{
\begin{tabular}{ll ccccc ccccc ccccc}
\toprule
& & \multicolumn{5}{c}{\textbf{DeepCAD}} & \multicolumn{5}{c}{\textbf{Fusion360}} & \multicolumn{5}{c}{\textbf{MCB}} \\
\cmidrule(lr){3-7} \cmidrule(lr){8-12} \cmidrule(lr){13-17}
Decoding & Model
& IoU\,\%$\uparrow$ & CD$_{8k}\downarrow$ & CD$_{30k}\downarrow$ & GMS\,\%$\uparrow$ & IR\,\%$\downarrow$
& IoU\,\%$\uparrow$ & CD$_{8k}\downarrow$ & CD$_{30k}\downarrow$ & GMS\,\%$\uparrow$ & IR\,\%$\downarrow$
& IoU\,\%$\uparrow$ & CD$_{8k}\downarrow$ & CD$_{30k}\downarrow$ & GMS\,\%$\uparrow$ & IR\,\%$\downarrow$ \\
\midrule
\multirow{2}{*}{Greedy}
& SFT & 91.7 & 0.16 & 0.048 & 95.3 & 2.63
      & 88.8 & 0.14 & 0.046 & 90.8 & 3.77
      & 75.2 & 0.43 & 0.259 & 63.3 & 12.04 \\
& RL  & 96.1 & 0.15 & 0.042 & 97.0 & 0.35
      & 94.1 & 0.12 & 0.038 & 93.3 & 1.22
      & 88.3 & 0.22 & 0.093 & 73.7 & 0.74 \\
\midrule
\multirow{2}{*}{Sampling}
& SFT & 93.6 & 0.15 & 0.046 & 95.6 & 0.15
      & 91.8 & 0.13 & 0.042 & 92.3 & 0.64
      & 82.8 & 0.31 & 0.150 & 70.0 & 0.36 \\
& RL  & \textbf{96.6} & \textbf{0.15} & \textbf{0.042} & \textbf{97.4} & \textbf{0.07}
      & \textbf{95.2} & \textbf{0.12} & \textbf{0.036} & \textbf{94.2} & \textbf{0.23}
      & \textbf{89.5} & \textbf{0.22} & \textbf{0.089} & \textbf{75.1} & \textbf{0.06} \\
\bottomrule
\end{tabular}}
\end{table*}

\section{Limitations}

Three limitations bound the present system. First, CADENA does not lead every part family: on gears \& bearings, CADFit's direct surface fitting reaches $61.3$ GMS against our $58.1$ (Table~\ref{tab:benchmark}), and a method that optimises geometry against the target does not depend on its training distribution covering rotationally patterned parts. Second, we do not yet report a single-pass model trained on our own corpus, which would separate the contribution of stepwise inference from that of the training data; we intend to include it in a later version. Third, CADENA emits CadQuery, and inherits its limits: in corner cases a construction tree that CadQuery executes has no faithful counterpart in industrial CAD software, so an exported program is not guaranteed to survive the transfer.

Beyond these, the failure modes in Appendix~\ref{sec:supp:failures} bound the method from outside the loop. The policy observes a fixed eight-view protocol, so geometry hidden from every canonical viewpoint is invisible to it; features no DSL operation expresses are approximated by stacks of extrusions; countable features such as gear teeth are approximated rather than counted, because one tooth too few leaves almost the same residual as the correct part; and an early operation that fixes the wrong plane is worked around by later steps rather than undone. Stepwise inference also executes and renders after every operation, which costs more per part than a single forward pass. 

\section{Conclusion}
\label{sec:conclusion}

CADENA outperforms prior methods on every dataset we evaluate, though the margin is far from uniform: the benefit of stepwise reconstruction grows with geometric complexity, leaving it modest on sketch--extrude corpora and substantially larger on real mechanical parts. Reinforcement learning against executed geometry strengthens both halves of that result, improving reconstruction accuracy while reducing the invalid rate. A different lesson comes from the comparison with CADFit, whose failures are complementary to our own. The two are better combined than treated as alternatives.

\bibliographystyle{unsrtnat}
\bibliography{references}

\clearpage
\appendix
\section{Supplementary Material}
\label{sec:supp}

This appendix reports the reward-variant experiments behind the RL design
(Section~\ref{app:rl-experiments}), the greedy-versus-sampling comparison
(Section~\ref{sec:decoding}), the surface grammar of the DSL
(Section~\ref{sec:supp:dsl}), and the recurring failure modes of the method
(Section~\ref{sec:supp:failures}).

\subsection{Comparison with SOV-CAD}
\label{sec:supp:sovcad}

SOV-CAD~\citep{sovcad} is the closest concurrent method, but it releases no
trained model, so we cannot run it under the protocol used everywhere else in
this paper. Table~\ref{tab:sovcad} therefore places the DeepCAD figures it
reports beside ours as published.

\begin{table}[h]
\centering
\caption{\textbf{CADENA against the DeepCAD figures reported by SOV-CAD.} Their
values are quoted from their Table~II (the stronger of their two variants); ours
are the CADENA-RL row of Table~\ref{tab:external}. The two are measured under
different protocols --- see the note below --- so this is a comparison of
published numbers, not a controlled one.}
\label{tab:sovcad}
\small
\setlength{\tabcolsep}{8pt}
\begin{tabular}{lcc}
\toprule
 & SOV-CAD & CADENA-RL \\
\midrule
Median CD$\downarrow$ & 0.38  & \textbf{0.15}  \\
IoU$\uparrow$         & 0.84  & \textbf{0.961} \\
IR$\downarrow$        & 7.3\% & \textbf{0.3\%} \\
\bottomrule
\end{tabular}
\tabnote{The gap is if anything understated. Our protocol evaluates in a fixed
frame rather than renormalising each prediction by its own bounding box, and
averages only over parts for which a metric is defined; both choices cost us
several points relative to the convention their numbers follow
(Section~\ref{sec:results}), so restating their results under our protocol would
widen the margin rather than narrow it. Their table does not state the point count behind its median Chamfer distance; we quote our $8$k value, so that row should be read as indicative and the comparison rested on IoU and IR, which are unambiguous.}
\end{table}

\subsection{DSL grammar and argument conventions}
\label{sec:supp:dsl}

This section documents the surface syntax the policy emits, as implemented by the
procedural generator that produces the training corpus. Every form below was
extracted from the generator and then checked against $994$ emitted programs
($15{,}084$ lines); operations with no program on disk were checked against
programs generated fresh from the repository's own presets. Forms that the source
can in principle emit but that occur in no generated program are omitted rather
than documented, since we cannot attest to them.

\paragraph{Program structure.} A program opens with a fixed import preamble and
the line \texttt{r = None}, after which each step is one line that consumes the
running solid \texttt{r} and rebinds it. The chain is what makes a step
well-formed after any prefix: because no line binds a name that a later line must
resolve, a step is a valid continuation of any earlier state, which is precisely
the property that lets the policy emit one operation from the observation alone.
Three constructs depart from the plain \texttt{r = op(r, \ldots)} shape and are
therefore worth stating explicitly:
\begin{itemize}\itemsep2pt
  \item \textbf{Auxiliary workplanes.} A standalone statement
        \texttt{w\{i\}=cq.Workplane('AXIS',origin=(x,y,z))} introduces a named
        workplane. It is the one construct that binds a name other than
        \texttt{r}, and it is consumed only by \texttt{gear}.
  \item \textbf{First-operation \texttt{gear}.} \texttt{gear} takes an auxiliary
        workplane rather than \texttt{r}: \texttt{r = gear(w0, \ldots)}. In the
        corpus it therefore appears only as a first operation.
  \item \textbf{Edge operations.} \texttt{fillet} and \texttt{chamfer} are
        emitted as CadQuery selector chains,
        \texttt{r=r.edges(PointOnEdgeSelector([x,\,y,\,z])).fillet(radius)},
        rather than as calls of the form \texttt{op(r,\ldots)}. They still consume
        and rebind \texttt{r}.
\end{itemize}

\paragraph{Shared arguments.} Most 3D operations share a leading
\texttt{(r, point, plane, sketch, \ldots)} prefix, with the following conventions.

\begin{itemize}\itemsep2pt
  \item \textbf{\texttt{point}} --- a 3-tuple in \emph{world} coordinates, not
        sketch-local, expressed in the normalised integer frame defined in
        Section~\ref{sec:method}. The component along the plane
        normal sets the workplane offset; when \texttt{r} is not \texttt{None} the
        point is additionally snapped onto the existing solid, and the in-plane
        components steer that snap rather than positioning the profile.
  \item \textbf{\texttt{plane}} --- a quoted enum with exactly three values,
        \texttt{'XY'}, \texttt{'YZ'}, \texttt{'ZX'}, with normals $+Z$, $+X$, $+Y$
        respectively. No other token occurs.
  \item \textbf{\texttt{sketch}} --- a double-quoted CadQuery sketch chain that
        always begins \texttt{sketch()} and ends \texttt{.finalize()}. Its
        coordinates are \emph{global in-plane} $(u,v)$ of the named plane, so the
        string carries no offset information. Observed vocabulary:
        \texttt{push}, \texttt{segment}, \texttt{arc}, \texttt{close},
        \texttt{assemble}, \texttt{circle}, \texttt{rect}, \texttt{reset},
        \texttt{face}, \texttt{wires}, and \texttt{mode='s'} for subtractive
        sub-profiles.
  \item \textbf{Lengths} --- signed, in the same normalised integer units as
        \texttt{point}; positive is along the positive plane normal.
\end{itemize}

\begin{table}[t]
\centering
\caption{\textbf{Operations of the CADENA DSL.} Surface form as emitted, grouped
by what the operation does to the running solid. Bracketed arguments are optional
and do occur in generated programs. \emph{Attested} is the number of occurrences
in the $994$-program corpus; the operation marked $\ast$ does not appear there and was
verified against programs generated fresh from the repository presets, and $\checkmark$ marks one that is
attested in the corpus but whose occurrences we did not tally. The generator
implements further operations that the CADENA corpus does not use; they are omitted.}
\label{tab:dsl}
\small
\setlength{\tabcolsep}{4pt}
\resizebox{\linewidth}{!}{
\begin{tabular}{llr}
\toprule
Operation & Emitted form & Attested \\
\midrule
\multicolumn{3}{l}{\emph{Sketch-based}} \\
extrude   & \texttt{r=extrude(r, pt, 'PL', "sk", h[, on\_surf])} & 1312 \\
revolve   & \texttt{r = revolve(r, pt, 'PL', "sk", angle, 'AXIS')} & 312 \\
hole      & \texttt{r = hole(r, pt, 'PL', "sk", depth)} & 470 \\
shell     & \texttt{r=shell(r, pt, 'PL', "sk", h, wall, bottom)} & \multicolumn{1}{r}{$\checkmark$} \\
orto\_cut & \texttt{r=orto\_cut(r, pt, 'PL', "sk", extent)} & 170 \\
\midrule
\multicolumn{3}{l}{\emph{Swept and lofted}} \\
loft      & \texttt{r=loft(r, pt, 'PL', [profile, \ldots][, ruled=True])} & 89 \\
sweep     & \texttt{r=sweep(r, "profile", "path")} & 278 \\
sweep\_adv & \texttt{r=sweep\_adv(r, "profile", "path")} & 110 \\
spring    & \texttt{r=spring(r, pt, 'PL', profile, pitch, height, radius,} & \multirow{2}{*}{$\ast$} \\
          & \hspace{1em}\texttt{angle, centre, seed[, body\_mode=, turns=, tails=])} & \\
\midrule
\multicolumn{3}{l}{\emph{Parametric features}} \\
gear      & \texttt{r = gear(w0, outer\_radius=, cylinder\_height=,} & \multirow{2}{*}{26} \\
          & \hspace{1em}\texttt{number\_outer\_teeth=, outer\_tooth\_profile=, \ldots)} & \\
\midrule
\multicolumn{3}{l}{\emph{Edge operations} (selector chains)} \\
fillet    & \texttt{r=r.edges(PointOnEdgeSelector([x,y,z])).fillet(rad)} & 180 \\
chamfer   & \texttt{r=r.edges(PointOnEdgeSelector([x,y,z])).chamfer(w[, w2])} & 177 \\
\bottomrule
\end{tabular}}
\end{table}

\paragraph{Naming note.} The helical-sweep operation is spelled \texttt{helix} in
material generated before the generator was refactored and \texttt{spring}
afterwards; the argument list is unchanged. 

\subsection{RL experiments}
\label{app:rl-experiments}

The row labelled RL (IoU) is the checkpoint reported as CADENA-RL in Table~\ref{tab:external}; SFT rows in this appendix are the final checkpoint of the second supervised stage. Where a value here disagrees with the main text in the last printed digit, the main-text rounding is authoritative.

The RL procedure --- one-step GRPO against the programmatic environment, a
self-generated reference trajectory, group filtering and candidate selection ---
is described in Section~\ref{sec:method}. This appendix varies one thing only,
the reward, and reports what each variant does to reconstruction quality. It
carries the evidence behind two statements made there: that the KL penalty does
not earn its place in the objective, and that GMS is used strictly for
evaluation and never as a training signal.

\subsubsection{Setup}
\label{app:setup}

All configurations below are trained on the RL dataset described in
Section~\ref{sec:method} and differ only in the reward function. Evaluation is
performed on the 1000-example subsamples \texttt{deepcad-1000},
\texttt{fusion360-1000} and \texttt{mcb-1000}, and, for the one pair where a
subsample difference had to be checked against noise, on the full test splits.
All numbers are obtained under greedy decoding.

\subsubsection{Reward definitions}
\label{app:rewards}

\paragraph{IoU.} As in the main text, the volumetric IoU between the constructed
solid and the target mesh, computed directly from the mesh. The RL (IoU)
configuration is trained on this reward using the base procedure, without a KL
penalty.

\paragraph{MPR (mean precision/recall).} $N = 20{,}000$ points are sampled from
the surfaces of the target and predicted meshes; for each point, the absolute
signed distance to the opposite mesh is computed. For a threshold $\tau$:

\begin{equation}
P(\tau) = \frac{1}{N}\bigl|\{p \in \hat{S} : d(p, S) \le \tau\}\bigr|, \qquad
R(\tau) = \frac{1}{N}\bigl|\{p \in S : d(p, \hat{S}) \le \tau\}\bigr|,
\end{equation}

\noindent where $S$ and $\hat{S}$ are the target and predicted surfaces,
respectively. The value at a given threshold is aggregated as a weighted sum
$M(\tau) = w_P P(\tau) + w_R R(\tau)$, and the final reward averages two
thresholds with fixed weights:

\begin{equation}
\mathrm{MPR} = 0.6 \, M(0.5) + 0.4 \, M(1.0).
\end{equation}

Distances are measured in the normalised frame of Section~\ref{sec:method}, with
the part's bounding box scaled so that its largest side equals 200 units. At
that scale $\tau = 0.5$ is roughly the manufacturing-tolerance range --- surface
deviations that are practically indistinguishable visually on a typically sized
CAD model --- while $\tau = 1.0$ additionally penalises coarser local shape
errors. Combining the two thresholds gives a two-level score: strict surface
closeness together with tolerance of larger, but still acceptable, deviations.
The RL (MPR, 0.5/0.5) configuration is trained with symmetric weights
$w_P = w_R = 0.5$; the RL (MPR, 0.9/0.1) configuration is biased toward
precision, with $w_P = 0.9$, $w_R = 0.1$.

\paragraph{IoU + KL.} The RL (IoU + KL) configuration is trained on the IoU
reward with an added KL penalty relative to the SFT reference policy.

\paragraph{GMS.} Whether the evaluation metric can double as a reward is tested
separately in Section~\ref{app:gms}.

\subsubsection{Comparison of reward variants}
\label{app:main-results}

\begin{table}[h]
\centering
\caption{Comparison of reward functions across the deepcad-1000,
fusion360-1000, and mcb-1000 datasets. All rows share the RL setup of
Section~\ref{sec:method} and differ only in the reward; decoding is greedy.}
\label{tab:main}
\resizebox{\textwidth}{!}{
\begin{tabular}{lcccc cccc cccc}
\toprule
& \multicolumn{4}{c}{deepcad-1000} & \multicolumn{4}{c}{fusion360-1000} & \multicolumn{4}{c}{mcb\_1000} \\
\cmidrule(lr){2-5}\cmidrule(lr){6-9}\cmidrule(lr){10-13}
Model & IoU$\uparrow$ & CD$\downarrow$ & GMS$\uparrow$ & IR$\downarrow$ & IoU$\uparrow$ & CD$\downarrow$ & GMS$\uparrow$ & IR$\downarrow$ & IoU$\uparrow$ & CD$\downarrow$ & GMS$\uparrow$ & IR$\downarrow$ \\
\midrule
SFT & 0.9190 & 0.1564 & 0.9549 & 0.0417 & 0.8795 & 0.1444 & 0.9062 & 0.0571 & 0.8465 & 0.1213 & 0.7713 & 0.1174 \\
RL (IoU) & 0.9620 & \textbf{0.1467} & 0.9737 & 0.0081 & 0.9317 & \textbf{0.1236} & 0.9349 & 0.0310 & 0.9143 & \textbf{0.0929} & 0.8310 & 0.0091 \\
RL (MPR, 0.5/0.5) & 0.9465 & 0.1495 & 0.9703 & \textbf{0.0020} & 0.9114 & 0.1312 & 0.9294 & 0.0060 & 0.9003 & 0.0941 & 0.8258 & 0.0070 \\
RL (MPR, 0.9/0.1) & 0.9546 & 0.1487 & 0.9694 & \textbf{0.0020} & 0.9296 & 0.1312 & 0.9289 & \textbf{0.0030} & 0.9241 & 0.0947 & 0.8247 & 0.0010 \\
RL (IoU + KL) & \textbf{0.9650} & 0.1476 & \textbf{0.9743} & 0.0030 & \textbf{0.9377} & 0.1259 & \textbf{0.9387} & 0.0100 & \textbf{0.9340} & 0.0936 & \textbf{0.8339} & \textbf{0.0000} \\
\bottomrule
\end{tabular}
}
\end{table}

Every RL configuration improves on the SFT model on all three subsamples
(Table~\ref{tab:main}): mean IoU rises, median CD falls, and the fraction of
invalid programs drops.

MPR is competitive with the IoU reward and on some cells better. The symmetric
variant ($w_P = w_R = 0.5$) trails the IoU reward on mean IoU on
\texttt{mcb-1000}; shifting the weight toward precision ($w_P = 0.9$) not only
closes most of that gap but overtakes the IoU reward on that dataset.

Adding a KL penalty to the IoU reward is ahead on the subsamples: against
RL (IoU) it wins mean IoU, GMS and IR on all three datasets, and loses median CD
on all three. To separate that from subsample noise, the RL (IoU) /
RL (IoU + KL) pair --- the same checkpoints as in Table~\ref{tab:main} --- was
recomputed on the full test splits (Table~\ref{tab:full}). There the difference
is small and inconsistent in direction across datasets and metrics, so the
advantage seen on the subsamples is not confirmed at full scale and appears to
fall within noise; this is why the reported model is trained with $\beta = 0$.
The MPR and GMS configurations were not recomputed on the full splits, and our
conclusions about them therefore remain limited to the 1000-example subsamples.

\begin{table}[h]
\centering
\caption{Full-test-split evaluation for the RL (IoU) / RL (IoU + KL) pair --- same
checkpoints as in Table~\ref{tab:main}. RL (MPR, 0.9/0.1) was not evaluated on
the full splits.}
\label{tab:full}
\resizebox{\textwidth}{!}{
\begin{tabular}{lcccc cccc cccc}
\toprule
& \multicolumn{4}{c}{deepcad\_test\_mesh} & \multicolumn{4}{c}{fusion360\_test\_mesh} & \multicolumn{4}{c}{mcb\_test\_mesh} \\
\cmidrule(lr){2-5}\cmidrule(lr){6-9}\cmidrule(lr){10-13}
Model & IoU$\uparrow$ & CD$_{8k}\downarrow$ & GMS$\uparrow$ & IR$\downarrow$ & IoU$\uparrow$ & CD$_{8k}\downarrow$ & GMS$\uparrow$ & IR$\downarrow$ & IoU$\uparrow$ & CD$_{8k}\downarrow$ & GMS$\uparrow$ & IR$\downarrow$ \\
\midrule
SFT           & 0.9167 & 0.1581 & 0.9528 & 0.0263 & 0.8880 & 0.1422 & 0.9079 & 0.0377 & 0.7524 & 0.4316 & 0.6339 & 0.1204 \\
RL (IoU)      & \textbf{0.9610} & 0.1477 & 0.9700 & \textbf{0.0035} & 0.9410 & 0.1243 & 0.9334 & 0.0122 & \textbf{0.8825} & \textbf{0.2236} & \textbf{0.7373} & \textbf{0.0074} \\
RL (IoU + KL) & 0.9589 & \textbf{0.1471} & \textbf{0.9723} & 0.0047 & \textbf{0.9427} & \textbf{0.1232} & \textbf{0.9377} & \textbf{0.0087} & 0.8821 & 0.2276 & 0.7361 & 0.0102 \\
\bottomrule
\end{tabular}
}
\end{table}

\subsubsection{GMS as a reward}
\label{app:gms}

We also tested whether GMS itself can be used as a reward. Both runs start from
the same intermediate SFT checkpoint under an otherwise identical setup, so the
two reward signals can be compared directly on the 1000-example subsamples.

\begin{table}[h]
\centering
\caption{GMS as a reward: the target metric (GMS) improves, but IoU and CD
degrade, especially on mcb-1000.}
\label{tab:gms}
\resizebox{\textwidth}{!}{
\begin{tabular}{lcccc cccc cccc}
\toprule
& \multicolumn{4}{c}{deepcad-1000} & \multicolumn{4}{c}{fusion360-1000} & \multicolumn{4}{c}{mcb\_1000} \\
\cmidrule(lr){2-5}\cmidrule(lr){6-9}\cmidrule(lr){10-13}
Model & IoU$\uparrow$ & CD$\downarrow$ & GMS$\uparrow$ & IR$\downarrow$ & IoU$\uparrow$ & CD$\downarrow$ & GMS$\uparrow$ & IR$\downarrow$ & IoU$\uparrow$ & CD$\downarrow$ & GMS$\uparrow$ & IR$\downarrow$ \\
\midrule
SFT (intermediate) & 0.9108 & 0.1606 & 0.9461 & 0.0215 & 0.8534 & 0.1639 & 0.8821 & 0.0560 & 0.8030 & 0.1490 & 0.6863 & 0.0493 \\
RL (IoU) & \textbf{0.9451} & \textbf{0.1538} & 0.9527 & \textbf{0.0091} & \textbf{0.9013} & \textbf{0.1457} & 0.8935 & \textbf{0.0384} & \textbf{0.8748} & \textbf{0.1220} & 0.7028 & \textbf{0.0030} \\
RL (GMS) & 0.8982 & 0.1621 & \textbf{0.9646} & 0.0121 & 0.8427 & 0.1696 & \textbf{0.9032} & 0.0460 & 0.7588 & 0.3035 & \textbf{0.7285} & 0.0204 \\
\bottomrule
\end{tabular}
}
\end{table}

The GMS reward produces the largest gain in mean GMS on all three datasets
(Table~\ref{tab:gms}), and it buys that gain with geometry: mean IoU ends
\emph{below} the checkpoint the run started from on all three subsamples, and on
\texttt{mcb-1000} median CD more than doubles, from $0.1490$ to $0.3035$. The
IoU reward on the same checkpoint moves both in the right direction there,
raising mean IoU and lowering median CD. Optimising GMS therefore improves the
target metric without improving --- and sometimes at the expense of --- actual
geometric fidelity. We conclude that GMS is not a valid RL training signal and
use it strictly as an evaluation metric.

\subsection{Inference-time ablations}
\label{sec:supp:ablations}

All ablations below are computed from the stepwise records of the reported CADENA-RL model: because every prefix is executed and scored during inference, we can replay the run under different decision rules without retraining or re-running the model.

\begin{figure*}[t]
    \centering
    \includegraphics[width=\textwidth]{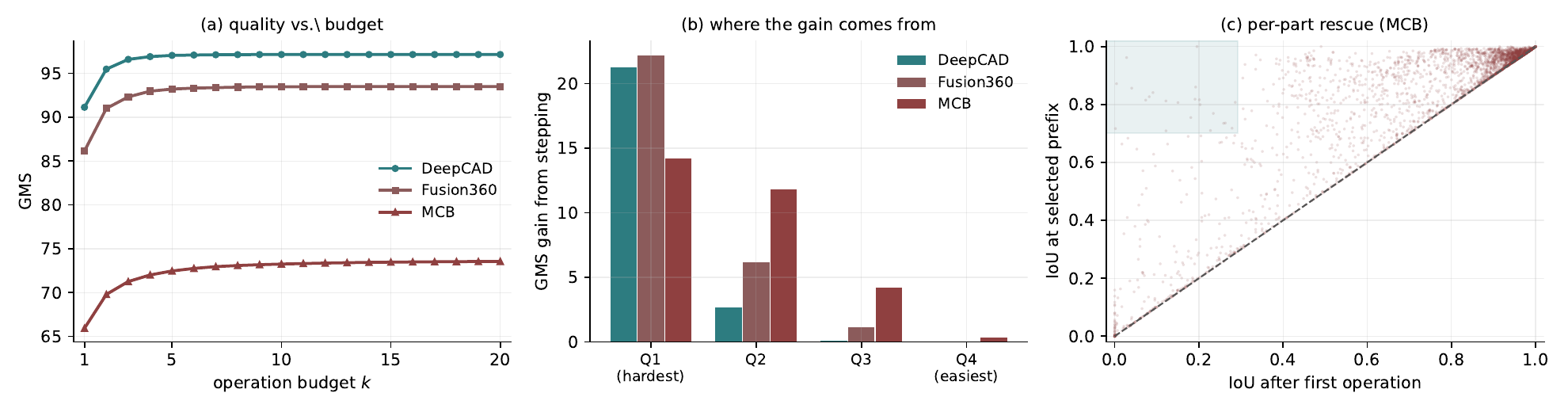}
    \caption{\textbf{Inference-time ablations}, all replayed from the stepwise records of the reported model. \textbf{(a)}~Quality against the operation budget $k$: $k=1$ is a single-pass model, and the curve saturates around eight operations, so the budget of 20 used throughout is not a tuned quantity. \textbf{(b)}~The gain is concentrated in the hardest quartile of each dataset --- parts whose first operation scores worst --- and is essentially zero for the easiest. \textbf{(c)}~Per-part view on MCB: each point is a part, comparing the reconstruction after one operation with the selected prefix. Points above the diagonal are improved by stepping; the shaded region marks parts rescued from near-total failure.}
    \label{fig:budget}
\end{figure*}

\paragraph{How much does stepping buy?} Restricting the policy to a budget of $k$ operations and returning its best prefix so far interpolates between a single-pass model ($k=1$) and the full method (Fig.~\ref{fig:budget}). The second operation alone is worth $+4.4$ GMS on DeepCAD, $+4.9$ on Fusion360, and $+3.9$ on MCB; the full budget adds $+6.0$, $+7.4$, and $+7.6$ respectively. Parts consume a median of 2 operations on DeepCAD and Fusion360 but 4 on MCB, so real mechanical parts do not merely score lower --- they require more of the loop.

\paragraph{Where does the gain come from?} Splitting each dataset into quartiles by the quality of the \emph{first} operation --- that is, by how well a single-pass answer would have done --- shows that the benefit is concentrated almost entirely in the hardest quartile (Table~\ref{tab:ablation_difficulty}). On DeepCAD the worst quartile gains $21.3$ GMS while the best gains nothing at all; on MCB, where more parts are hard, the two hardest quartiles gain $14.2$ and $11.9$. Stepwise generation is thus not a uniform improvement but a targeted one: it rescues the parts a single-pass model gets wrong, and leaves the easy ones alone.

\begin{table}[t]
\centering
\caption{\textbf{GMS gain from the operation budget, by difficulty quartile.} Q1 contains the parts whose first operation scores worst. Gains concentrate where single-pass generation fails.}
\label{tab:ablation_difficulty}
\small
\begin{tabular}{lcccc}
\toprule
& \multicolumn{2}{c}{$k=1$ (single-pass)} & \multicolumn{2}{c}{$k=20$ (full)} \\
\cmidrule(lr){2-3}\cmidrule(lr){4-5}
Quartile & DeepCAD & MCB & DeepCAD & MCB \\
\midrule
Q1 (hardest) & 72.8 & 39.2 & 94.1 \gain{+21.3} & 53.4 \gain{+14.2} \\
Q2           & 95.0 & 58.9 & 97.7 \gain{+2.7}  & 70.8 \gain{+11.9} \\
Q3           & 98.1 & 78.3 & 98.2 \gain{+0.1}  & 82.5 \gain{+4.2} \\
Q4 (easiest) & 98.7 & 87.3 & 98.7 \gain{+0.0}  & 87.7 \gain{+0.4} \\
\bottomrule
\end{tabular}
\end{table}

\paragraph{Which parts are rescued?} Comparing the first operation against the selected prefix per part, $14\%$ of DeepCAD parts, $17\%$ of Fusion360 parts, and $15\%$ of MCB parts improve by at least $0.10$ IoU, and a smaller group is rescued outright from near-total failure (below $0.30$ IoU at the first step to above $0.70$ at the selected prefix): 66, 24, and 22 parts respectively. The remainder are already solved by their first operation and are left unchanged, which is the desired behaviour.

\paragraph{Does prefix selection matter?} Returning the best prefix rather than the last is worth $+0.27$ GMS on DeepCAD, $+0.40$ on Fusion360, and $+0.30$ on MCB, at no cost. Selecting instead by GMS --- the metric we report --- would add only a further $+0.18$, $+0.30$, and $+0.57$, confirming that the IoU-based rule is not implicitly optimising the evaluation metric.

\paragraph{Does the model follow its corpus or the geometry?} Section~\ref{sec:intro} argued that a single-pass model inherits the compositional biases of its training programs, whereas conditioning on the residual frees the choice of operation from position in the sequence. This is testable. We parse the operation type of every step, both in the training corpus (40k sampled steps) and in the model's own output on each test set, and compare the distributions (Table~\ref{tab:composition}).

The corpus is close to position-agnostic: \texttt{extrude} accounts for 34.8--53.6\% of steps at every depth, and the distribution at position 5 looks much like the one at position 2. A model that merely replayed these statistics would therefore produce roughly the same operation mix on any input. CADENA does not. On DeepCAD it opens with \texttt{extrude} in 80.1\% of parts, against 53.6\% in the corpus; on MCB --- whose parts are dominated by turned, rotationally symmetric bodies --- it opens with \texttt{revolve} in 73.2\% of parts, against 32.5\% in the corpus, and \texttt{extrude} falls to 26.0\%. The same model, trained on the same programs, inverts its leading operation according to the geometry in front of it. From the second operation onward it concentrates on \texttt{hole} (60.4--84.2\%, against 20.0--21.1\% in the corpus), the natural consequence of having built a body that must now be cut.

We read this as direct evidence for the claim: the operation distribution CADENA produces is a function of the shape being reconstructed rather than of the corpus prior, which is precisely what a single-pass decoder --- with no access to the intermediate geometry --- cannot condition on.

\begin{table}[t]
\centering
\caption{\textbf{Operation distribution, corpus vs.\ inference} (\% of steps). The training corpus is nearly position-agnostic, while CADENA's choice of first operation inverts between DeepCAD and MCB according to the geometry.}
\label{tab:composition}
\small
\begin{tabular}{llcccc}
\toprule
Position & Operation & Training corpus & DeepCAD & Fusion360 & MCB \\
\midrule
\multirow{2}{*}{1st} & extrude & 53.6 & 80.1 & 70.3 & 26.0 \\
                     & revolve & 32.5 & 16.1 & 27.2 & 73.2 \\
\midrule
\multirow{2}{*}{2nd} & extrude & 34.8 & 36.6 & 38.9 & 31.1 \\
                     & hole    & 20.0 & 62.9 & 60.4 & 67.6 \\
\midrule
\multirow{2}{*}{5th} & extrude & 46.8 & 20.4 & 23.5 & 15.3 \\
                     & hole    & 21.1 & 79.3 & 75.7 & 84.2 \\
\bottomrule
\end{tabular}
\end{table}

\subsection{Full evaluation on CADENA-Bench}
\label{sec:supp:fulleval}

We additionally report a full breakdown of all evaluation metrics for every model on CADENA-Bench, both in aggregate and separately for each of the six part-family classes. Table~\ref{tab:cadenabench_all} reports the aggregate results over all $n=3396$ parts, and Tables~\ref{tab:cadenabench_71_72_73}--\ref{tab:cadenabench_74_75_76} report the corresponding results for classes, respectively. The metrics are provided as in the main text; CD, IoU, and GMS are computed only over the samples for which a model produced a valid prediction. CADFit$^\dagger$ denotes the CADFit baseline evaluated under the same fixed-compute-budget convergence criterion described in the main text, which similarly biases its reported means toward an easier subset of samples.

This distinction between IR and the remaining metrics is important for a fair reading of the tables: a model can appear strong on CD, IoU, or GMS simply by being evaluated only on the subset of well-formed, easier cases it successfully handled, while silently failing on the rest. For example, cadrille attains the lowest IR across almost all classes but does so together with substantially worse CD, IoU, and GMS than our RL model, indicating a conservative failure mode rather than genuinely higher-quality reconstructions. Conversely, CADReasoner and CADFit$^\dagger$ occasionally post competitive CD or IoU on individual classes (e.g., class Flat \& levers) alongside comparatively high IR, and CADFit$^\dagger$ collapses outright on class Tooling \& gauges (IR $=23.5\%$, CD$_{8k}=6.81$), showing that strong numbers on a subset of metrics do not necessarily generalize to the full distribution of geometries.

Taking IR into account, CADENA-RL consistently achieves the best trade-off between reconstruction quality and reliability. It obtains the best CD$_{8k}$, CD$_{30k}$, IoU, and GMS both in the overall comparison (Table~\ref{tab:cadenabench_all}) and in nearly every individual class (Tables~\ref{tab:cadenabench_71_72_73}--\ref{tab:cadenabench_74_75_76}), with the sole exception of GMS on class Gears \& bearings, where CADFit$^\dagger$ is marginally higher despite its much higher IR on that class. At the same time, CADENA-RL maintains one of the lowest IR values among all models in every class, trailing only cadrille, which achieves a lower IR at the cost of substantially weaker geometric fidelity throughout. Together, these results indicate that the gains of CADENA-RL are not an artifact of evaluating on an easier subset of samples, but reflect genuinely more accurate and more reliable reconstructions across the full range of part complexities represented in CADENA-Bench.

\begin{table}[t]
\centering
\caption{\textbf{CADENA-Bench, all metrics over the full benchmark} ($n=3396$). Median CD at 8k and 30k sampled points, mean IoU and GMS, and the invalid rate. CD, IoU and GMS are computed only over parts for which a method returned a valid prediction, so each row's means describe a different subset and IR is the column that says how large that subset is. This is the invalid rate for CADENA-Bench that Table~\ref{tab:benchmark} does not carry.}
\label{tab:cadenabench_all}
\begin{tabular}{lccccc}
\toprule
Method & CD$_{8k}\downarrow$ & CD$_{30k}\downarrow$ & IoU$\uparrow$ & GMS$\uparrow$ & IR$\downarrow$ \\
\midrule
CAD-Recode  & 1.0614 & 0.9048 & 68.64 & 48.89 & 27.9 \\
cadrille    & 1.2673 & 1.1253 & 68.77 & 49.79 & \textbf{0.8} \\
CADReasoner & 1.9707 & 1.8327 & 68.34 & 52.87 & 39.9 \\
CADEvolve   & 1.3117 & 1.1433 & 73.00 & 52.90 & 10.8 \\
CADFit$^\dagger$ & \underline{0.7125} & \underline{0.5364} & \underline{75.69} & 54.80 & 18.8 \\
\midrule\midrule
CADENA-SFT  & 0.9643 & 0.7738 & 73.88 & \underline{57.96} & 12.8 \\
CADENA-RL   & \textbf{0.3310} & \textbf{0.1763} & \textbf{87.57} & \textbf{66.95} & \underline{1.6} \\
\bottomrule
\end{tabular}
\end{table}

\begin{table}[t]
\centering
\caption{\textbf{CADENA-Bench by part family: shafts \& bushings, gears \& bearings, housings \& frames.} Metrics and averaging rule as in Table~\ref{tab:cadenabench_all}. Gears \& bearings is the family on which the learned methods lose most, and the only one where direct fitting outscores CADENA.}
\label{tab:cadenabench_71_72_73}
\resizebox{\textwidth}{!}{
\begin{tabular}{l ccccc ccccc ccccc}
\toprule
& \multicolumn{5}{c}{Shafts \& bushings ($n=762$)} & \multicolumn{5}{c}{Gears \& bearings ($n=749$)} & \multicolumn{5}{c}{Housing \& frames ($n=359$)} \\
\cmidrule(lr){2-6}\cmidrule(lr){7-11}\cmidrule(lr){12-16}
Method & CD$_{8k}$ & CD$_{30k}$ & IoU & GMS & IR & CD$_{8k}$ & CD$_{30k}$ & IoU & GMS & IR & CD$_{8k}$ & CD$_{30k}$ & IoU & GMS & IR \\
\midrule
CAD-Recode  & 1.0089 & 0.8402 & 71.63 & 53.75 & 20.1 & 1.2859 & 1.0795 & 62.34 & 39.09 & 35.9 & \underline{1.1932} & \underline{1.0762} & 73.50 & 56.08 & 34.3 \\
cadrille    & 1.0506 & 0.9071 & 73.29 & 57.55 & \textbf{0.9} & 1.4828 & 1.2861 & 65.72 & 37.18 & \textbf{0.4} & 1.8209 & 1.7479 & 70.12 & 54.46 & \textbf{0.6} \\
CADReasoner & 2.0882 & 1.9453 & 69.41 & 55.55 & 21.0 & 2.5867 & 2.4641 & 65.69 & 46.90 & 74.2 & 2.1856 & 2.0205 & 71.72 & 56.41 & 39.8 \\
CADEvolve   & 0.9855 & 0.8518 & 77.68 & 59.57 & 10.9 & 1.6314 & 1.4526 & 67.78 & 41.34 & 12.1 & 1.6275 & 1.4723 & 75.58 & \underline{59.39} & 8.6 \\
CADFit$^\dagger$ & 0.4536 & 0.2711 & 81.53 & 60.81 & 16.1 & \underline{0.5592} & \underline{0.3590} & \underline{79.61} & \textbf{61.27} & 18.6 & 1.7366 & 1.5505 & 63.48 & 48.45 & 25.6 \\
\midrule\midrule
CADENA-SFT  & \underline{0.3176} & \underline{0.1415} & \underline{86.36} & \underline{70.76} & 11.7 & 1.8563 & 1.6688 & 63.07 & 47.75 & 11.6 & 1.9574 & 1.8238 & \underline{77.00} & 58.06 & 11.4 \\
CADENA-RL   & \textbf{0.2825} & \textbf{0.1030} & \textbf{91.77} & \textbf{73.03} & \underline{1.3} & \textbf{0.4907} & \textbf{0.3043} & \textbf{85.03} & \underline{58.13} & \underline{0.8} & \textbf{0.5807} & \textbf{0.4449} & \textbf{87.22} & \textbf{70.74} & \underline{1.9} \\
\bottomrule
\end{tabular}
}
\end{table}

\begin{table}[t]
\centering
\caption{\textbf{CADENA-Bench by part family: flat \& levers, springs \& fasteners, tooling \& gauges.} Metrics and averaging rule as in Table~\ref{tab:cadenabench_all}. Tooling \& gauges holds 17 parts, too few for its column to be read as a ranking; it is reported for completeness.}
\label{tab:cadenabench_74_75_76}
\resizebox{\textwidth}{!}{
\begin{tabular}{l ccccc ccccc ccccc}
\toprule
& \multicolumn{5}{c}{Flat \& levers ($n=464$)} & \multicolumn{5}{c}{Springs \& fasteners ($n=1046$)} & \multicolumn{5}{c}{Tooling \& gauges ($n=17$)} \\
\cmidrule(lr){2-6}\cmidrule(lr){7-11}\cmidrule(lr){12-16}
Method & CD$_{8k}$ & CD$_{30k}$ & IoU & GMS & IR & CD$_{8k}$ & CD$_{30k}$ & IoU & GMS & IR & CD$_{8k}$ & CD$_{30k}$ & IoU & GMS & IR \\
\midrule
CAD-Recode  & 0.3931 & 0.3175 & 72.96 & 58.79 & 33.5 & \underline{1.1215} & 0.9922 & 67.17 & 45.31 & 23.4 & 1.8759 & 1.7713 & 67.27 & 38.66 & 23.5 \\
cadrille    & 0.3682 & 0.2645 & 70.82 & 62.06 & \textbf{0.9} & 1.4312 & 1.2847 & 66.03 & 46.31 & \textbf{1.0} & 2.1844 & 2.0251 & 75.48 & 41.79 & \textbf{0.0} \\
CADReasoner & 0.2964 & 0.1945 & 74.57 & \underline{69.67} & 29.4 & 2.8767 & 2.7127 & 62.75 & 43.36 & 33.7 & 1.8975 & 1.7782 & \underline{82.11} & 38.47 & 47.1 \\
CADEvolve   & 0.3448 & 0.2345 & 76.26 & 66.40 & 8.9 & 1.4773 & 1.3425 & \underline{71.00} & 47.89 & 11.4 & 2.0745 & 1.9259 & 75.45 & \underline{44.67} & 17.6 \\
CADFit$^\dagger$ & 0.2585 & 0.1709 & \underline{79.92} & 66.20 & 17.7 & 1.5809 & 1.4122 & 66.34 & 42.97 & 18.9 & 6.8112 & 6.5517 & 40.12 & 23.05 & 23.5 \\
\midrule\midrule
CADENA-SFT  & \underline{0.2453} & \underline{0.1670} & 75.79 & 66.75 & 10.4 & 1.1688 & \underline{0.9830} & 70.85 & \underline{51.92} & 16.2 & \underline{1.4242} & \underline{1.3335} & 67.87 & 43.36 & \underline{5.9} \\
CADENA-RL   & \textbf{0.1586} & \textbf{0.0566} & \textbf{88.38} & \textbf{76.81} & \underline{2.4} & \textbf{0.3432} & \textbf{0.1986} & \textbf{85.94} & \textbf{63.35} & \underline{1.8} & \textbf{0.3290} & \textbf{0.1903} & \textbf{91.49} & \textbf{59.98} & \underline{5.9} \\
\bottomrule
\end{tabular}
}
\end{table}

\subsection{Methods we could not compare against}
\label{sec:supp:omissions}
 The tables above omit several recent systems that target the same setting. Each is absent for a different reason, and we state them explicitly because a silent omission is indistinguishable from an unfavourable one. SOV-CAD~\citep{sovcad} is the closest in formulation --- likewise stepwise, likewise driven by rendered views --- but its repository ships evaluation code only, listing pretrained checkpoints and model definitions as still to come, so it cannot be executed at all. CADFS~\citep{cadfs} releases code and weights, but emits FeatureScript, Onshape's proprietary language: recovering geometry from its predictions requires an Onshape account with per-team API keys and a round trip through their hosted kernel for every part, which is not practical at the scale of these test sets and would make our numbers depend on a closed, rate-limited service. Zero-to-CAD~\citep{zerotocad} releases weights but no renderer, and specifies its input only as eight $256\times256$ views, ``four front-facing and four rear-facing''. We recovered that convention from the released renders --- orthographic isometric at $\arctan(1/\sqrt{2})$ elevation, four azimuths per hemisphere, shaded with black feature edges --- and validated it end to end on the authors' own test split, where our renders reach 55.6 IoU against 68.0 obtained with the authors' own images. At 82\% of that ceiling a tabulated number would confound their method with our reconstruction of their renderer, so we omit it. For HistCAD~\citep{histcad} and IterCAD~\citep{itercad} we found no public implementation. Running a method through an approximated or proprietary input pipeline understates it in a way the reader cannot detect, which is why these are reported as omissions rather than as weak rows.

\subsection{Failure cases}
\label{sec:supp:failures}

\begin{figure*}[t]
    \centering
    \includegraphics[width=\linewidth]{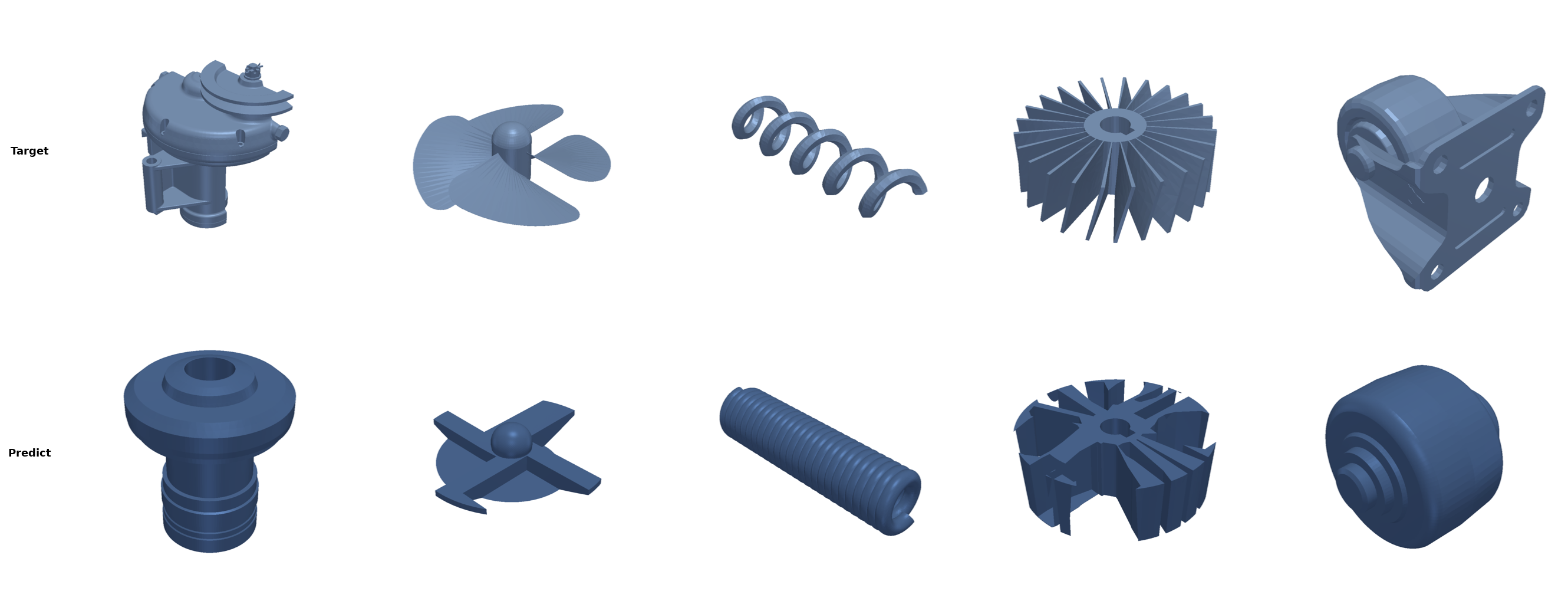}
    \caption{\textbf{Characteristic failures.} Five parts on which CADENA fails, with the target above and the returned reconstruction below. They illustrate three of the four modes discussed in this section: countable features are approximated rather than counted, so the impeller and the radial fan keep the right body and the wrong blades; helical geometry outside the DSL's vocabulary is replaced by the nearest available primitive, a thread standing in for a coil spring; and where no single canonical view resolves the shape, the reconstruction collapses to a rough envelope of it, as in the valve body and the bracket. None of these is a failure to emit valid CadQuery --- every reconstruction shown builds and is watertight, and would be counted in the means of Table~\ref{tab:external}.}
\label{fig:failures}
\end{figure*}

Four failure modes recur (Fig.~\ref{fig:failures}), and they are worth separating
because only the first is a failure of the \emph{loop} rather than of its inputs
or its vocabulary.

\paragraph{Countable features are approximated rather than counted.} On gears and
splined shafts the reconstruction reproduces the body and the general character of
the teeth while getting their number wrong. The residual after the body is built
is a thin, repetitive shell that changes little whether the count is right or off
by one, so the signal that would correct the count is weak exactly where it is
needed. This is the largest contributor to the gap on gears \& bearings, our
weakest family on CADENA-Bench (Table~\ref{tab:benchmark}).

\paragraph{Occluded geometry is invisible to the observation.} The policy sees eight views --- six
axis-aligned and two isometric. Internal cavities, blind bores and undercuts that project
identically to a solid region are not represented in the observation at all, so no
amount of stepping recovers them --- the loop converges happily to a solid whose
exterior matches. This is a limitation of the render protocol rather than of
stepwise generation, and it is the motivation for adding a point-cloud modality.

\paragraph{Out-of-vocabulary features are approximated by what the DSL has.}
Geometry that no operation in Table~\ref{tab:dsl} can express --- freeform
blends, non-circular helical profiles, draft --- is approximated by stacks of
extrusions. The result is often close in Chamfer distance and clearly wrong to an
engineer, which is precisely the discrepancy GMS is designed to expose.

\paragraph{Early frame errors are not recovered.} When the first operation fixes a
plane or an origin inconsistent with the target, later steps can add material that
reduces the residual locally without ever undoing the original error. Because
selection returns the best prefix rather than the last, such a part is not made
worse by continued stepping, but neither is it repaired: the budget is spent
improving a reconstruction built on the wrong frame.

\end{document}